%% file: neurips_2026.tex
\documentclass{article}

\PassOptionsToPackage{numbers, sort}{natbib}
\usepackage[preprint]{neurips_2026}

\usepackage[utf8]{inputenc} 
\usepackage[T1]{fontenc}    
\usepackage{hyperref}       
\usepackage{url}            
\usepackage{booktabs}       
\usepackage{amsfonts}       
\usepackage{nicefrac}       
\usepackage{microtype}      
\usepackage[table]{xcolor}         

\usepackage[numbers]{natbib}

\input{preamble}

\title{\ourshortname: Test-time Baking for Efficient Rendering of Decoder-only View Synthesis Models}

\author{%
  Chin-Yang Lin$^{1,2}$\thanks{This work was conducted during an internship at NVIDIA.}\hspace{0.8em}%
  Ryo Hachiuma$^{1}$\hspace{0.8em}%
  Jaesung Choe$^{1}$\hspace{0.8em}%
  Min-Hung Chen$^{1}$ \\
  \bfseries Yu-Chiang Frank Wang$^{1}$\hspace{0.8em}%
  Yen-Yu Lin$^{2}$\hspace{0.8em}%
  Wei-Chen Chiu$^{2}$\hspace{0.8em}%
  Yu-Lun Liu$^{2}$\hspace{0.8em}%
  Cheng Sun$^{1}$ \\[0.2em]
  $^1$NVIDIA \quad $^2$Department of Computer Science, National Yang Ming Chiao Tung University \\
}

\begin{document}


\maketitle

\input{fig/teaser}
\input{sec/0_abstract}

\input{fig/motivation}

\input{sec/1_intro}

\input{sec/2_related}

\input{fig/pipeline}

\input{sec/3_0_approach}

\input{sec/3_1_overview}

\input{sec/3_2_DVSM}

\input{sec/3_3_0_TTB}

\input{sec/3_3_1_repurpose}
\input{sec/3_3_2_ctx_pair}

\input{sec/3_3_3_arch_details}

\input{sec/3_3_4_infer_strategy}
\input{fig/qual_dl3dv}
\input{tab/re10k_dl3dv}

\input{fig/qual_wild}

\input{tab/in_the_wild}

\input{tab/ablation_ttb}

\input{tab/ablation_combine}
\input{sec/4_0_experiments}

\input{sec/4_1_details}

\input{sec/4_2_comparisons}

\input{sec/4_3_abla}
\input{sec/5_conclusion}

\begin{ack}
This work was supported by NVIDIA Taiwan AI Research \& Development Center (TRDC). This research was funded by the National Science and Technology Council, Taiwan, under Grants NSTC 112-2222-E-A49-004-MY2, 113-2628-E-A49-023-, 115-2628-E-A49-024-, and 111-2628-E-A49-018-MY4. Yu-Lun Liu acknowledges the Yushan Young Fellow Program by the MOE in Taiwan.
\end{ack}

{
\small
\bibliographystyle{plainnat}
\bibliography{main}
}



\newpage 

\appendix
\begin{center}
    {\Large\bf Supplementary material}
\end{center}

\paragraph{Overview.}
This supplementary material expands on the main paper with implementation details (Sec.~\ref{supp:common}--\ref{supp:localdup}), a method-level comparison with related decoder-only LVSM and TTT-based methods (Sec.~\ref{supp:discussion}), additional quantitative results (Sec.~\ref{supp:add_results}), and additional qualitative comparisons (Sec.~\ref{supp:qualitative}).
We further provide turntable rendering videos and an interactive HTML viewer in the supplementary attachment.

\section{Additional details}
\subsection{Common implementation details}
\label{supp:common}

Unless otherwise specified, all of our final models share the configuration described in this section.
The decoder consists of 12 blocks with a latent dimension of 768, organized as illustrated in main paper Fig.~\ref{fig:pipeline}.
Optimization is carried out with AdamW~\cite{adamw} using a weight decay of 0.05 and momentum coefficients $(0.9, 0.95)$.
The learning rate is annealed along a cosine schedule preceded by 2.5k linear warm-up iterations, peaking at $4{\times}10^{-4}$ during the lowest-resolution stage and dropping to $1{\times}10^{-5}$ when finetuning at higher resolutions.
The perceptual term in our rendering loss is weighted by $\lambda=0.2$.
Training runs on 64 A100 GPUs, and we cast the learnable layers to BFloat16 mixed precision during forward propagation.

Camera information is encoded with the Pl\"ucker parameterization~\cite{plucker}: each pixel is associated with a 6-dimensional vector $(\mathbf{r}_o\times\mathbf{r}_d,\,\mathbf{r}_d)$, where $\mathbf{r}_o$ denotes the camera origin, $\mathbf{r}_d$ the unit direction of the corresponding ray, and $\times$ the vector cross product.
We further normalize the context-view poses following the convention of GS-LRM~\cite{gslrm}.

Inside each transformer block, we adopt the pre-norm layout combined with QK-normalization.
Each attention layer uses 12 heads, and every fully-connected sublayer expands the hidden dimension by a factor of $4$.
All linear projections in our architecture are bias-free.

\subsection{TTB layer details}
\label{supp:ttb}
\input{sec/X_pseudocode}

This section elaborates on TTB layer details that were elided from the main paper for space.
We provide the pseudocode of training our TTB model in Algo.~\ref{algo:ttb_training} as a complementary reference for the descriptions below.
The three fast-weight matrices of our SwiGLU-MLP, $\mathbf{w}_0, \mathbf{w}_2 \in \mathbb{R}^{H \times d_\text{in} \times d_h}$ and $\mathbf{w}_1 \in \mathbb{R}^{H \times d_h \times d_\text{out}}$, are stored as ordinary learnable parameters and are jointly optimized in the outer loop alongside the rest of the network.
Consequently, the inner loop for every scene starts from a common, outer-loop-learned meta-initialization $\theta_0$, which is then advanced by $T$ Muon~\cite{muon} update steps before being applied to the rendering tokens.
When local-duplicate inference is enabled with $K>1$ groups, every per-group reconstruction starts
from the same meta-initialized $\theta_0$ and is specialized only by the views assigned to its
group, so all local copies share the same starting point and differ through their inner-loop
trajectories.

We initialize TTB models from the corresponding DVSM~\cite{dvsm} checkpoint with \texttt{strict=False}: the patch embeddings, input LayerNorms, Pl\"ucker encoding, intra-view attention layers, both FFNs in each block, and the RGB head are inherited directly, while every module inside the TTB layer—including its LayerNorm, query/key and value projections, output projection, per-token learning-rate head, output RMSNorm, and the fast weights $\mathbf{w}_0, \mathbf{w}_1, \mathbf{w}_2$—is freshly initialized.
The cross-view attention projections of DVSM are not reused, since the inner-loop target of TTB is structurally different and warm-starting from the DVSM projections did not yield benefits in our preliminary experiments.
To prevent the random-init TTB layer from disrupting the inherited DVSM features at the start of finetuning, we initialize the TTB output projection with a small standard deviation following the residual zero-init trick used in deep transformers, so that the TTB block contributes approximately zero at finetune step $0$ and the entire forward pass reduces to the pretrained DVSM model plus an identity TTB layer.
This zero-init together with the parameter inheritance above accounts for the $+1.02$ PSNR gain reported in Tab. ~\ref{tab:ablation_pretrain}.

For the TTB layer micro-architecture, we share a single linear projection between query and key construction: the inner-loop input $\mathbf{W}_k \mathbf{X}^{(\text{ctx-rend})}$ in main paper Eq.~\ref{eq:ttt_ttb_upd} and the rendering-time query $\mathbf{W}_k \mathbf{X}^{(\text{nv})}$ are both produced by passing the input through a single linear layer followed by SiLU and an $L_2$ unit-norm, which constrains the writes and reads of the fast-weight MLP to a common spherical subspace and removes a magnitude degree of freedom from the inner-loop optimization.
The per-token learning rate is predicted by an additional linear head that outputs $3H$ scalars per token—one for each of the three fast-weight matrices in every head—and reparameterized as $\mathrm{softplus}(\cdot + \mathrm{softplus}^{-1}(0.01))$, ensuring strictly positive learning rates and an effective initial value of $0.01$ across all heads and matrices.

The outer loop differentiates fully through all $T$ inner-loop steps without any first-order approximation, so that the rendering loss reflects the exact dependence of the final fast weights on the outer-loop parameters.
Each Muon step orthogonalizes the inner-loop gradient with five Newton--Schulz quintic iterations executed in BFloat16; we transpose the gradient before iteration whenever $d_\text{in} > d_h$ so that the intermediate Gram matrix has size $\min(d_\text{in}, d_h)^2$, which keeps the per-step memory manageable.
Even with this optimization, retaining $T$ inner-step activations for backpropagation remains the dominant memory cost of TTB.
We mitigate it with two layers of gradient checkpointing: the inner-loop activations of SwiGLU and Muon are recomputed within each step, and the outer-loop checkpointing is applied selectively to the intra-view attention and FFN sublayers but not to the TTB sublayer itself, the combination of which makes our 64-A100 training feasible at $960$p resolution with $V\!=\!32$ context views.

Finally, our two main paper variants differ only in the inner-loop mapping target.
TTB instantiates $g(\cdot)$ in main paper Eq.~\ref{eq:ttb_upd} as the cross-view attention output of $\mathbf{X}^{(\text{ctx-gt})}$, while TTB-- replaces this with a linear projection $\mathbf{W}_v \mathbf{X}^{(\text{ctx-gt})}$ acting directly on the context ground-truth tokens.
Both variants share every other component described above, including the context pair construction, the QK-shared projection, the per-token learning-rate head, the residual target, and the feedback path that propagates the baked MLP output back to $\mathbf{X}^{(\text{ctx-gt})}$.

\subsection{Two-view Re10K specific details}
\label{supp:re10k}

For the two-view setting, each GPU receives 32 scenes per iteration, with 2 context views and 3 target views per scene.
The interval between the two context frames is sampled uniformly between $25$ and $192$, and target frames are drawn from the union of all intermediate frames between the two context views and the extrapolated frames within $25$ frames on either side.
We initialize the model from the corresponding DVSM~\cite{dvsm} Re10K checkpoint following the strategy detailed in Sec.~\ref{supp:ttb}, and finetune it on Re10K for $100$K iterations at $256{\times}256$ followed by $10$K iterations at $512{\times}512$.
Note that doubling the context tokens through the context pair (main paper Eq.~\ref{eq:ctx_pair}) negligibly affects this two-view setup, so we keep the same per-GPU batch composition as DVSM.

\subsection{Many-view DL3DV~\cite{dl3dv} specific details}
\label{supp:dl3dv}

DL3DV is the main benchmark for our dense-view results in main paper Tab.~\ref{tab:dl3dv} and Tab.~\ref{tab:in_the_wild}, as well as for all the ablations in main paper Sec.~\ref{ssec:abla}.
For training, we adopt a dynamic batching scheme that controls the per-GPU token budget rather than the number of scenes.
Each scene draws either $16$ or $32$ context views uniformly at random, paired with the same number of target views; the number of scenes packed into a GPU is then adjusted on the fly so that the aggregate count of \emph{context-pair} tokens (i.e., the union of $\mathbf{X}^{(\text{ctx-gt})}$ and $\mathbf{X}^{(\text{ctx-rend})}$ across all scenes in the batch) stays within a fixed budget.
This compensates for the doubled token count introduced by the context pair and keeps the effective compute per iteration comparable to DVSM at the same memory ceiling.
The frame interval between two adjacent context views is sampled uniformly from $[1, 16]$, and target views are drawn from $[\min(\mathcal{C}){-}2, \max(\mathcal{C}){+}2] \setminus \mathcal{C}$, where $\mathcal{C}$ is the index set of the sampled context views.
We exclude from training a small subset of scenes whose aspect ratio or native resolution does not match the canonical $16{:}9$ $960$p layout.
The resolution curriculum starts from $340$p and runs for $100$K iterations, after which the model is finetuned for $10$K iterations at each of $480$p, $720$p, and the final $960$p ($540{\times}960$).

For the controlled ablation setting used in main paper Tab.~\ref{tab:ablation_ttb}--\ref{tab:ablation_feedback_residual} and Tab.~\ref{tab:ablation_pretrain}, we shrink the compute footprint to $8$ A100 GPUs by fixing the context view count to $32$, raising the patch size to $16$ to lower the per-token resolution, and truncating the curriculum to $340$p / $480$p / $720$p with $100$K / $10$K / $10$K iterations respectively.
All the design knobs ablated in those tables share this controlled setup; the rendering and reconstruction efficiency reported alongside is measured on a model with patch size $8$ to remain consistent with the final main-table configuration.

\subsection{Zero-shot evaluation details}
\label{supp:zeroshot}

The in-the-wild numbers in main paper Tab.~\ref{tab:in_the_wild} are produced by the DVSM-initialized $960$p TTB model trained on DL3DV with patch size $8$, evaluated zero-shot on three out-of-distribution datasets: the $9$ scenes of MipNeRF360~\cite{mip-nerf360}, the $7$ scenes of Free~\cite{f2nerf}, and the $6$ stable scenes from Hike~\cite{hike} as identified by~\cite{longsplat}.
We hold out every $8$th frame on MipNeRF360 and Free and every $10$th frame on Hike as target views, and run k-means over the remaining frames to obtain context views.
All input images are first resized—preserving their original aspect ratio—so that the total pixel count matches the $540{\cdot}960{=}518{,}400$ target used by our model, and predicted novel views are then resized back to the dataset's native resolution prior to metric computation.
We apply the same image preprocessing pipeline to all baselines for a fair comparison.
The LagerNVS~\cite{largernvs} numbers are produced from our intermediate $480$p checkpoint, since LagerNVS operates at $512$ pixels along the longer image side and matching its native input resolution is the cleanest comparison.

\subsection{Local-duplicate inference details}
\label{supp:localdup}

This section gives further details of the local-duplicate inference described in main paper
Sec.~\ref{sssec:infer_strategy}, which reconstructs the scene as $K$ spatially-local groups of
context views at inference time, each with its own copy of the baked fast weights, so that a
single fast-weight MLP only has to absorb the views of one local group (about $B$ on average)
instead of all $V$ views. It reuses the trained model as is, with no retraining and no additional
learned parameters; the per-scene state grows to $K$ sets of fast weights.

\paragraph{Grouping.}
Each context view $i$ is summarized into a 6-dimensional descriptor
$\mathbf{f}_i=[\mathbf{c}_i;\mathbf{d}_i]\in\mathbb{R}^6$, where $\mathbf{c}_i\in\mathbb{R}^3$ is
the camera origin and $\mathbf{d}_i\in\mathbb{R}^3$ is the camera's unit optical axis, both
expressed in the input world frame. The number of groups $K$ is an inference-time hyper-parameter;
$K=1$ falls back to the single-MLP model. Unless stated otherwise we set $K=V/B$ with a nominal
budget of $B=16$ views per group, \emph{i.e.}, $K=2$ for $V=32$ and $K=4$ for $V=64$. We then run
Lloyd's algorithm for 20 iterations with the Euclidean distance to cluster the $V$ descriptors into
$K$ groups. Centroids are initialized with the descriptors of $K$ views whose indices are evenly
spaced over $[0,V{-}1]$; since our context views are ordered by capture time, this spreads the
initial centroids along the camera trajectory. If a centroid attracts no views during an iteration,
we leave it unchanged for that round. Group sizes are neither balanced nor capped, and $K$ must
stay small relative to $V$.

\paragraph{Per-group reconstruction.}
For every group $k$ we run the standard reconstruction pass of our model on its member views only.
Inside this pass, all sublayers---intra-view attention, FFN, cross-view attention, and the TTB
inner loop---see only the tokens of group $k$; in particular, the cross-view attention that
provides the baking target attends across the views of the group rather than across all $V$ views.
The two streams $X^{(\text{ctx-gt})}$ and $X^{(\text{ctx-rend})}$ of a context view are created
inside the same pass and are therefore always baked together. Every pass starts from the shared
meta-initialization $\theta_0$ and runs the same $T$ inner-loop steps, producing the per-layer fast
weights $\{\theta_{k,\ell}\}_{\ell=1}^{L}$ of copy $k$. The $K$ passes are independent of each
other and are executed sequentially in our implementation. At the end of context prefilling we
cache the fast weights of every group together with the $K$ cluster centroids.

\paragraph{Rendering.}
At rendering time, we do not re-run k-means. Each novel view computes its own 6-dimensional
descriptor in the same input world frame and takes an $\arg\min$ of the Euclidean distance to the
cached centroids to select a single group, and all of its tokens are rendered with the fast weights
of that group. The switch between copies is hard, without blending. The per-token rendering cost is
unchanged and independent of $V$; routing only adds one nearest-centroid search over $K$ centroids
per view, which is negligible. In our implementation every novel view is routed and rendered
individually.

\section{Discussion: comparison with related methods}
\label{supp:discussion}

This section positions TTB within the recent landscape of feed-forward novel-view synthesis (NVS) and 3D reconstruction methods.
We focus on the LVSM family~\cite{lvsm,dvsm,efflvsm,lact} together with three recent methods that adopt either TTT layers (LongLRM~\cite{longlrm}, tttLRM~\cite{tttlrm}) or a 3D-pretrained encoder (LagerNVS~\cite{largernvs}), since these are the closest design points to TTB.
The discussion is structured around three axes: training settings (Sec.~\ref{supp:disc_train}), reconstruction/rendering complexity (Sec.~\ref{supp:disc_complexity}), and architectural composition (Sec.~\ref{supp:disc_arch}).

\subsection{Training settings across methods}
\label{supp:disc_train}

Tab.~\ref{tab:disc_train} compares the training data, compute, and pretraining source across baselines and our method.
All listed methods share AdamW as the outer-loop optimizer and a multi-stage resolution curriculum, so we restrict the table to the three axes that most directly affect comparability.

\begin{table}[h]
\centering
\caption{\textbf{Training settings across methods.} ``Multi-dataset'' indicates training on a curated mixture beyond a single benchmark. ``DVSM ckpt'' refers to initializing from the corresponding DVSM~\cite{dvsm} checkpoint released for that benchmark. ``--'' marks compute resources that are not explicitly reported by the corresponding paper.}
\label{tab:disc_train}
\small
\begin{tabular}{l l l l}
\toprule
Method & Training data & GPUs & Pretrain source \\
\midrule
LVSM~\cite{lvsm} & Objaverse / Re10K & 64$\times$A100 & from scratch \\
Efficient-LVSM~\cite{efflvsm} & Re10K & 64$\times$A100 & from scratch \\
DVSM~\cite{dvsm} & DL3DV / Re10K & 64$\times$A100 & from scratch \\
LaCT~\cite{lact} & Objaverse / DL3DV & 64$\times$A100 & from scratch \\
LongLRM~\cite{longlrm} & DL3DV & 64$\times$A100 & from scratch \\
tttLRM~\cite{tttlrm} & Objaverse / DL3DV & 64$\times$A100 & LaCT~\cite{lact} ckpt \\
LagerNVS~\cite{largernvs} & Multi-dataset & --- & VGGT~\cite{vggt} ckpt \\
\midrule
TTB (ours) & DL3DV / Re10K & 64$\times$A100 & DVSM~\cite{dvsm} ckpt \\
\bottomrule
\end{tabular}
\end{table}

The compute footprint ($64{\times}$A100) is broadly shared across the group when reported, so the differences mainly lie in data composition and pretraining strategy.
Most LVSM-family methods train scene-only on Re10K or DL3DV, occasionally extending to Objaverse for object-level pretraining; LagerNVS is the outlier that explicitly relies on a multi-dataset mixture of fifteen sources to obtain its in-the-wild generalization.
Three methods escape the from-scratch pattern, and they do so along two distinct lines: tttLRM and TTB warm-start from an NVS checkpoint of the same task family (TTT-LVSM and DVSM, respectively), reusing weights that were already optimized for novel-view rendering; LagerNVS instead transfers from VGGT, a model trained for 3D reconstruction, and jointly finetunes both the VGGT encoder and a freshly initialized cross-attention renderer with the rendering loss.
TTB sits closer to tttLRM in this taxonomy—both inherit cross-view-attention features from a pretrained NVS backbone and reshape them into a TTT-style fast-weight representation—while LagerNVS represents an orthogonal direction of leveraging 3D-supervised priors.

\subsection{Reconstruction and rendering complexity}
\label{supp:disc_complexity}

Tab.~\ref{tab:disc_complexity} summarizes per-scene reconstruction and per-token rendering complexity for the methods above.
The scene representation column reports the asymptotic memory footprint of the per-scene state that is consumed at rendering time.

\begin{table}[h]
\centering
\caption{\textbf{Asymptotic complexity comparison.} 
$V$ denotes the number of context views and $L$ the number of tokens per view.
Reconstruction is reported per scene, rendering per novel-view token. ``GS rasterizer'' denotes that rendering is handled by a Gaussian splatting rasterizer rather than a neural module, which is not directly comparable in $\mathcal{O}(\cdot)$ form.
We only list the complexity of the main model layer for global scene understanding.
LongLRM still employs a few cross-view attention layers with $\mathcal{O}(V^2 \cdot L^2)$ complexity.
LaCT, tttLRM, LagerNVS, and ours also employ intra-view attention with $\mathcal{O}(V \cdot L^2)$ reconstruction complexity and $\mathcal{O}(L)$ per-token rendering complexity.
}
\label{tab:disc_complexity}
\small
\begin{tabular}{l l l l}
\toprule
Method & Reconstruction & Rendering (per token) & Scene representation \\
\midrule
LVSM (decoder-only)~\cite{lvsm} & --- & $\mathcal{O}(V \cdot L)$ & context tokens \\
LVSM (enc.-dec.)~\cite{lvsm} & $\mathcal{O}(V^2 \cdot L^2)$ & $\mathcal{O}(L_s)$ & latent scene tokens \\
Efficient-LVSM~\cite{efflvsm} & $\mathcal{O}(V \cdot L^2)$ & $\mathcal{O}(V \cdot L)$ & KV-cache \\
DVSM~\cite{dvsm} & $\mathcal{O}(V^2 \cdot L^2)$ & $\mathcal{O}(V \cdot L)$ & KV-cache \\
LaCT~\cite{lact} & $\mathcal{O}(V \cdot L)$ & $\mathcal{O}(1)$ & fast weights \\
LongLRM~\cite{longlrm} & $\mathcal{O}(V \cdot L)$ & GS rasterizer & 3D Gaussians \\
tttLRM~\cite{tttlrm} & $\mathcal{O}(V \cdot L)$ & GS rasterizer & 3D Gaussians \\
LagerNVS~\cite{largernvs} & $\mathcal{O}(V^2 \cdot L^2)$ & $\mathcal{O}(V \cdot L)$ & latent scene tokens \\
\midrule
TTB{-}{-} (ours) & $\mathcal{O}(V \cdot L)$ & $\mathcal{O}(1)$ & fast weights \\
TTB (ours) & $\mathcal{O}(V^2 \cdot L^2)$ & $\mathcal{O}(1)$ & fast weights \\
\bottomrule
\end{tabular}
\end{table}

The table makes the design space explicit.
The decoder-only LVSM reconstructs nothing offline and pays the full attention cost at every novel-view query; LVSM enc.-dec.\ and Efficient-LVSM amortize the cost into a one-time encoding stage but the rendering cost remains linear in the number of context tokens.
DVSM further removes the encoder-decoder decoupling but at the price of quadratic cross-view attention during prefilling.
LagerNVS occupies the same asymptotic regime as LVSM enc.-dec.\ but with a heavier $V$-dependence because its VGGT encoder runs alternating frame-wise and global self-attention over all $V \cdot L$ tokens, yielding $\mathcal{O}(V^2 \cdot L^2)$ encoding and $\mathcal{O}(V \cdot L)$ per-token rendering through cross-attention to the encoder output.
LaCT pushes both axes to their constant-complexity limits via TTT layers, whereas TTB occupies a previously empty cell on this map: it preserves DVSM's quadratic reconstruction cost—necessary because cross-view attention provides the inner-loop target—while inheriting LaCT's $\mathcal{O}(1)$ rendering and fixed-size scene representation.
TTB-- in turn shows that, when the cross-view attention target is replaced by a linear projection, reconstruction collapses back to linear in $V$, recovering LaCT's complexity profile but with the rest of the TTB design choices retained.
LRM-style baselines such as LongLRM and tttLRM occupy a different regime altogether, since their per-scene state is an explicit 3D Gaussian set rendered by a rasterizer rather than a neural module; we list them for completeness but do not directly compare $\mathcal{O}$-rendering with them.

\subsection{Architectural composition}
\label{supp:disc_arch}

Tab.~\ref{tab:disc_arch} compares the per-block composition and the total decoder depth.
We restrict the table to depth, latent dim, and the cross-view sublayer type, since these are the most consequential differences relative to TTB.
For LagerNVS we report depth in two parts since its model is composed of a $24$-block VGGT encoder operating at $1024$-dim and a $12$-block cross-attention decoder operating at $768$-dim.

\begin{table}[h]
\centering
\caption{\textbf{Architectural composition.} ``Cross-view sublayer'' indicates the module that exchanges information across views in each block. The depth column is the number of decoder blocks, with the encoder depth listed in parentheses for encoder-decoder methods. The latent-dim column reports the decoder width, with the encoder width also given for encoder-decoder methods.}
\label{tab:disc_arch}
\small
\setlength{\tabcolsep}{4pt}
\begin{tabular}{@{}l c c p{0.42\linewidth}@{}}
\toprule
Method & \# blocks & Latent dim & Cross-view sublayer \\
\midrule
LVSM~\cite{lvsm} & 24 & 768 & cross-view attention \\
Efficient-LVSM~\cite{efflvsm} & 24 & 1024 & cached KV cross-attention \\
DVSM~\cite{dvsm} & 12 & 768 & cross-view attention (cached KV) \\
LaCT~\cite{lact} & 24 & 768 & large-chunk TTT layer \\
LongLRM~\cite{longlrm} & 24 & 768 & Mamba2 + attention hybrid \\
tttLRM~\cite{tttlrm} & 24 & 768 & large-chunk TTT layer \\
LagerNVS~\cite{largernvs} & 12 ($+$24 enc.) & 768 ($+$1024 enc.) & VGGT alt.\ frame/global attn.\ (enc.)\ + cross-attn.\ (dec.) \\
\midrule
TTB{-}{-} (ours) & 12 & 768 & TTB layer (linear target) \\
TTB (ours) & 12 & 768 & cross-view attn.\ + TTB layer \\
\bottomrule
\end{tabular}
\end{table}

A consistent pattern across the methods is a $24$-block decoder for from-scratch models, while methods that initialize from a pretrained backbone are able to reduce decoder depth.
DVSM and our method use a $12$-block decoder by sharing weights between reconstruction and rendering stages; LagerNVS reaches the same $12$-block decoder width and depth ($768$, $12$ blocks) by offloading representational capacity to a $24$-block, $1024$-dim VGGT encoder.
TTB inherits the $12$-block depth directly from DVSM: every TTB layer additionally houses a fast-weight MLP whose inner loop is supervised by the cross-view attention output, but the surrounding intra-view attention and FFN sublayers are unchanged from DVSM.
The net effect is that TTB matches or surpasses both $24$-block from-scratch baselines and the much heavier LagerNVS on DL3DV (main paper Tab.~\ref{tab:dl3dv}, Tab.~\ref{tab:in_the_wild}) while operating with comparable parameter counts to DVSM ($244$M for TTB-- and $271$M for TTB, against $163$M for DVSM, $379$M for LaCT, and over $1$B for LagerNVS due to the included VGGT encoder).
We attribute this efficiency to the combination of a shallow shared decoder and the inheritance of pretrained DVSM weights, which Sec.~\ref{supp:ttb} discusses in detail.

\input{sec/X_re10k_full}
\input{tab/re10k_detail}

\section{Qualitative results}
\label{supp:qualitative}

This section provides additional qualitative results that complement the quantitative comparisons in the main paper.
We include in-domain comparisons on DL3DV~\cite{dl3dv} (Sec.~\ref{supp:qual_dl3dv}), two-view comparisons on Re10K~\cite{stereomag} (Sec.~\ref{supp:qual_re10k}), and zero-shot comparisons on three in-the-wild datasets~\cite{mip-nerf360,f2nerf,hike} (Sec.~\ref{supp:qual_inthewild}).

\subsection{In-domain results on DL3DV}
\label{supp:qual_dl3dv}

\begin{figure}[p]
    \centering
    \includegraphics[width=\linewidth]{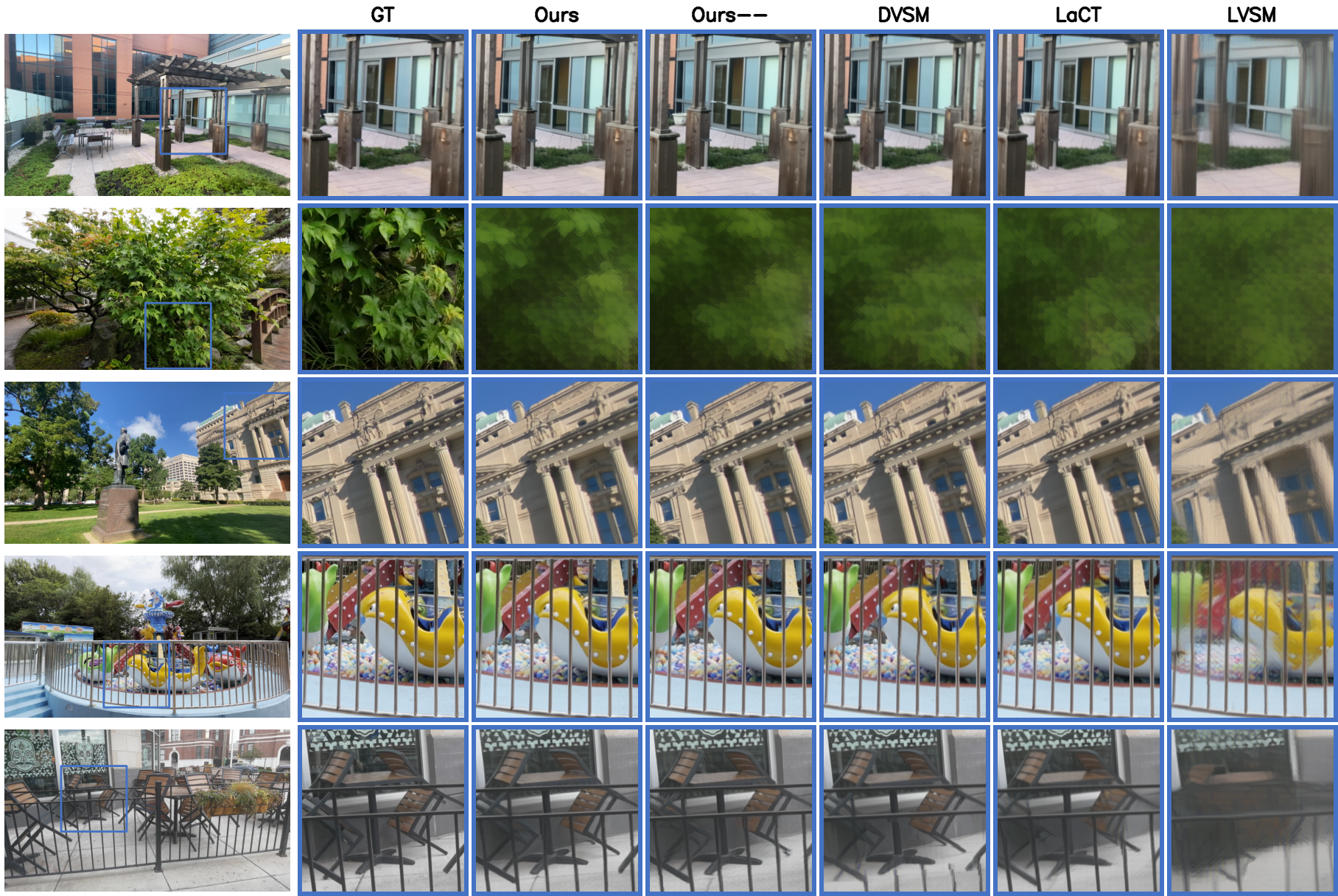}
    \caption{\textbf{Additional qualitative comparison on DL3DV.} 64 context views.}
    \label{fig:qual_dl3dv_more}
\end{figure}

Fig.~\ref{fig:qual_dl3dv_more} extends the comparison in main paper Fig.~\ref{fig:qual_dl3dv} to additional held-out DL3DV scenes spanning input view counts of $V{=}16$, $32$, and $64$, complementing the three-column quantitative results in main paper Tab.~\ref{tab:dl3dv}.
TTB consistently preserves thin geometric structures and high-frequency texture where LaCT~\cite{lact} blurs and LVSM~\cite{lvsm} loses detail.

\subsection{Two-view results on Re10K}
\label{supp:qual_re10k}

\begin{figure}[p]
    \centering
    \includegraphics[width=0.9\linewidth]{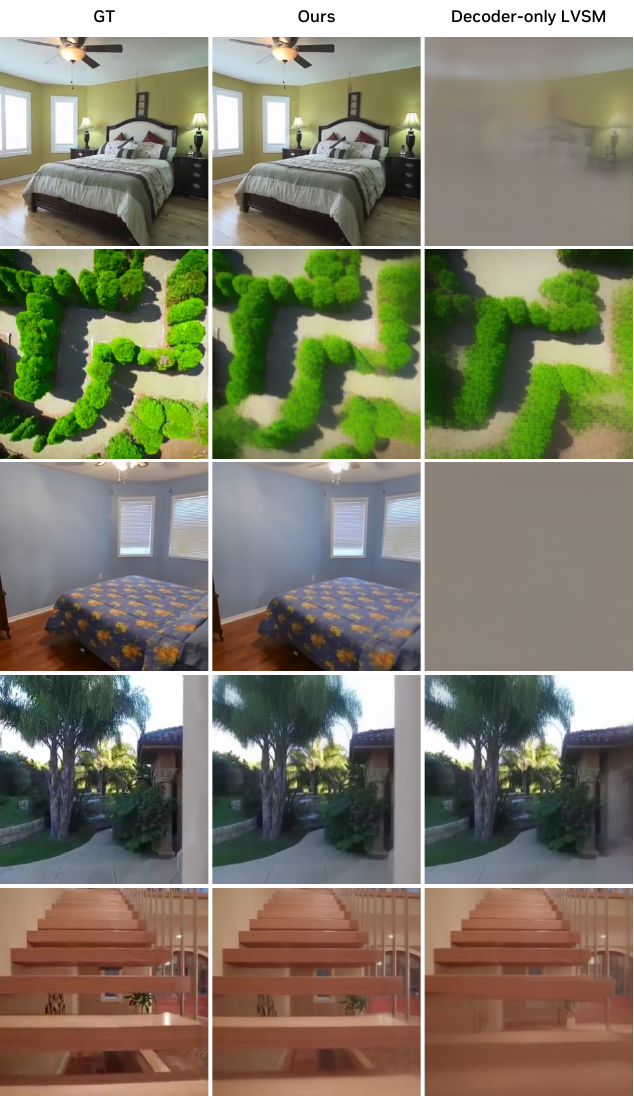}
    \caption{\textbf{Qualitative comparison on Re10K.} Two-view setting with sparse input.}
    \label{fig:qual_re10k}
\end{figure}

Fig.~\ref{fig:qual_re10k} compares TTB against DVSM~\cite{dvsm} and LVSM~\cite{lvsm} under the two-view setting, complementing main paper Tab.~\ref{tab:re10k}.
TTB matches DVSM's quality at the same rendering speed-up regime, and outperforms LVSM in regions involving moderate viewpoint extrapolation.

\subsection{Zero-shot results on in-the-wild datasets}
\label{supp:qual_inthewild}

\begin{figure}[p]
    \centering
    \includegraphics[width=\linewidth]{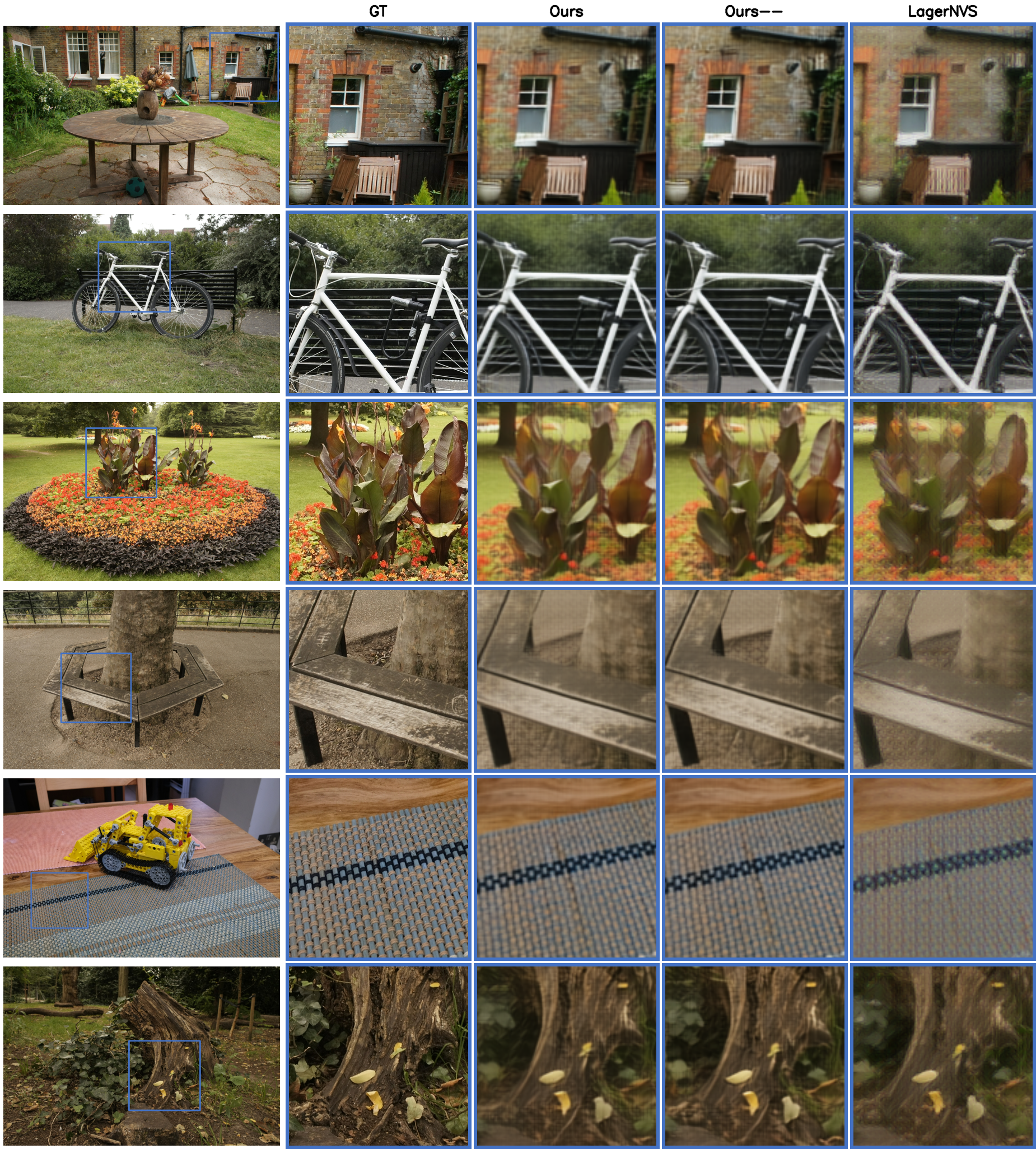}
    \caption{\textbf{Zero-shot qualitative results on MipNeRF360.} 64 context views.}
    \label{fig:qual_mipnerf360}
\end{figure}

\begin{figure}[p]
    \centering
    \includegraphics[width=\linewidth]{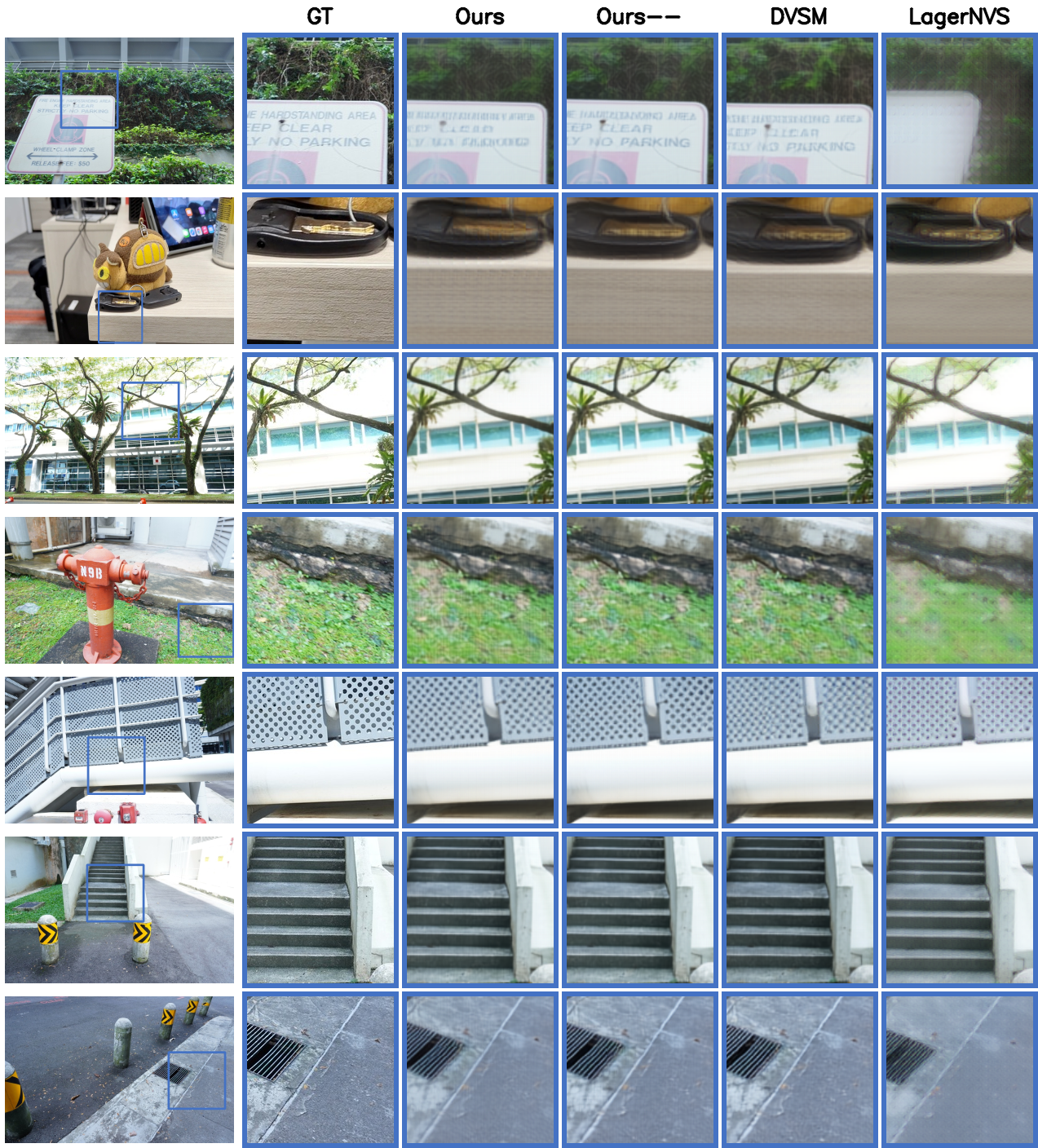}
    \caption{\textbf{Zero-shot qualitative results on Free.} 64 context views.}
    \label{fig:qual_free}
\end{figure}

\begin{figure}[p]
    \centering
    \includegraphics[width=\linewidth]{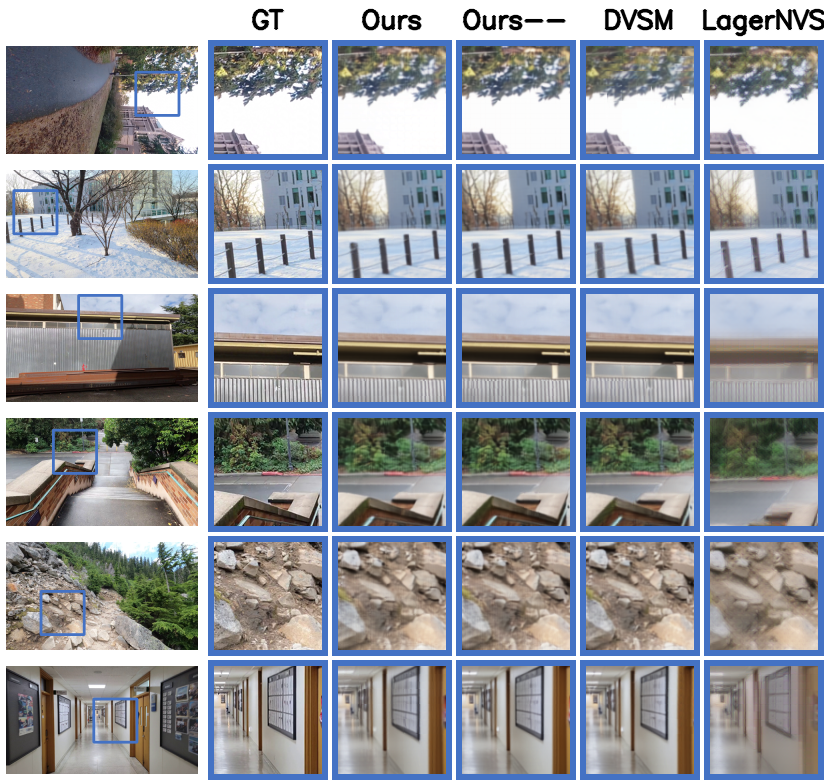}
    \caption{\textbf{Zero-shot qualitative results on Hike.} 64 context views.}
    \label{fig:qual_hike}
\end{figure}

Figs.~\ref{fig:qual_mipnerf360}--\ref{fig:qual_hike} report zero-shot results on MipNeRF360~\cite{mip-nerf360}, Free~\cite{f2nerf}, and Hike~\cite{hike}, all using the DL3DV-trained TTB model with $V{=}64$ input views.
These figures complement the in-the-wild numerical results in main paper Tab.~\ref{tab:in_the_wild}.
Despite a clear domain gap from DL3DV, TTB recovers fine details such as small text and thin structures that LaCT~\cite{lact} and LVSM~\cite{lvsm} blur out.

\section{Broader impacts}
\label{supp:impact}

TTB shares its broader impacts with feed-forward novel-view synthesis methods in general.
What is more specific to TTB is the methodological perspective it offers: rather than replacing attention with a test-time-trained fast-weight MLP as in prior TTT-style architectures, TTB shows that the fast-weight MLP can be supervised to \emph{collaborate} with attention by baking its outputs into a constant-size state.
This baking perspective is potentially applicable beyond novel-view synthesis, in any setting where a powerful but expensive attention module needs to be amortized into a state of fixed size.
Long-context language models, where attention KV-caches grow linearly with sequence length, could similarly benefit from baking attention outputs into a fixed-capacity fast weight to enable constant-memory inference.
World models and video generators that maintain a long temporal context face the same memory-growth issue and may adopt analogous baking strategies for the temporal-attention component.
Physical 3D perception systems that operate on streaming sensor inputs, such as embodied agents and autonomous driving stacks, could likewise use a baked fast-weight state to maintain a compact running memory of the environment without unbounded KV growth.
We hope the baking perspective inspires further exploration of how attention and test-time training can be combined across these domains.

\section{Declaration of LLM usage}
\label{supp:llm}

Large language models were used as writing and editing aids during manuscript preparation—paraphrasing, grammar correction, and improving readability.
We additionally used Claude Code as a coding assistant during the development of our training and evaluation pipeline; all code generated with its help was reviewed and tested by the authors before being incorporated into the final implementation.
LLMs did not contribute to the research ideation, model design, experimental protocol, or analysis of results.

\clearpage
\newpage

\end{document}

%% file: preamble.tex
\usepackage{multirow}
\usepackage{adjustbox}
\usepackage{xfrac}
\usepackage{array}
\usepackage{wrapfig}
\usepackage{colortbl}
\usepackage{xspace}

\usepackage{algorithm}
\usepackage{algpseudocode}
\usepackage{amsmath}
\usepackage{amssymb}

\usepackage{comment}

\newcommand{\ourshortname}{TTB\xspace}

\definecolor{downred}{RGB}{248,131,121}
\definecolor{upgreen}{RGB}{25,182,163}

\newcommand{\Real}{\mathbb{R}}

\usepackage{pifont}
\newcommand{\cmark}{\ding{51}}
\newcommand{\xmark}{\ding{55}}

\newcommand{\Image}{\mathbf{I}}
\newcommand{\Pose}{\mathbf{P}}
\newcommand{\Intr}{\mathbf{K}}

\newcommand{\rend}{\mathrm{Rend}}
\newcommand{\recon}{\mathrm{Recon}}

\newcommand{\PErgb}{\mathrm{PE}^{\text{(rgb)}}}
\newcommand{\PEray}{\mathrm{PE}^{\text{(ray)}}}
\newcommand{\token}{\mathbf{X}}

\newcommand{\tokenctx}{\token^{\text{(ctx)}}}
\newcommand{\tokenctxgt}{\token^{\text{(ctx-gt)}}}
\newcommand{\tokenctxrend}{\token^{\text{(ctx-rend)}}}
\newcommand{\tokennv}{\token^{\text{(nv)}}}

\newcommand{\attnintra}{\mathrm{IntraAttn}}
\newcommand{\attncross}{\mathrm{CrossAttn}}
\newcommand{\ffn}{\mathrm{FFN}}

%% file: fig/teaser.tex
\begin{figure}[h]
    \centering
    \vspace{-10mm}
    \includegraphics[width=\linewidth]{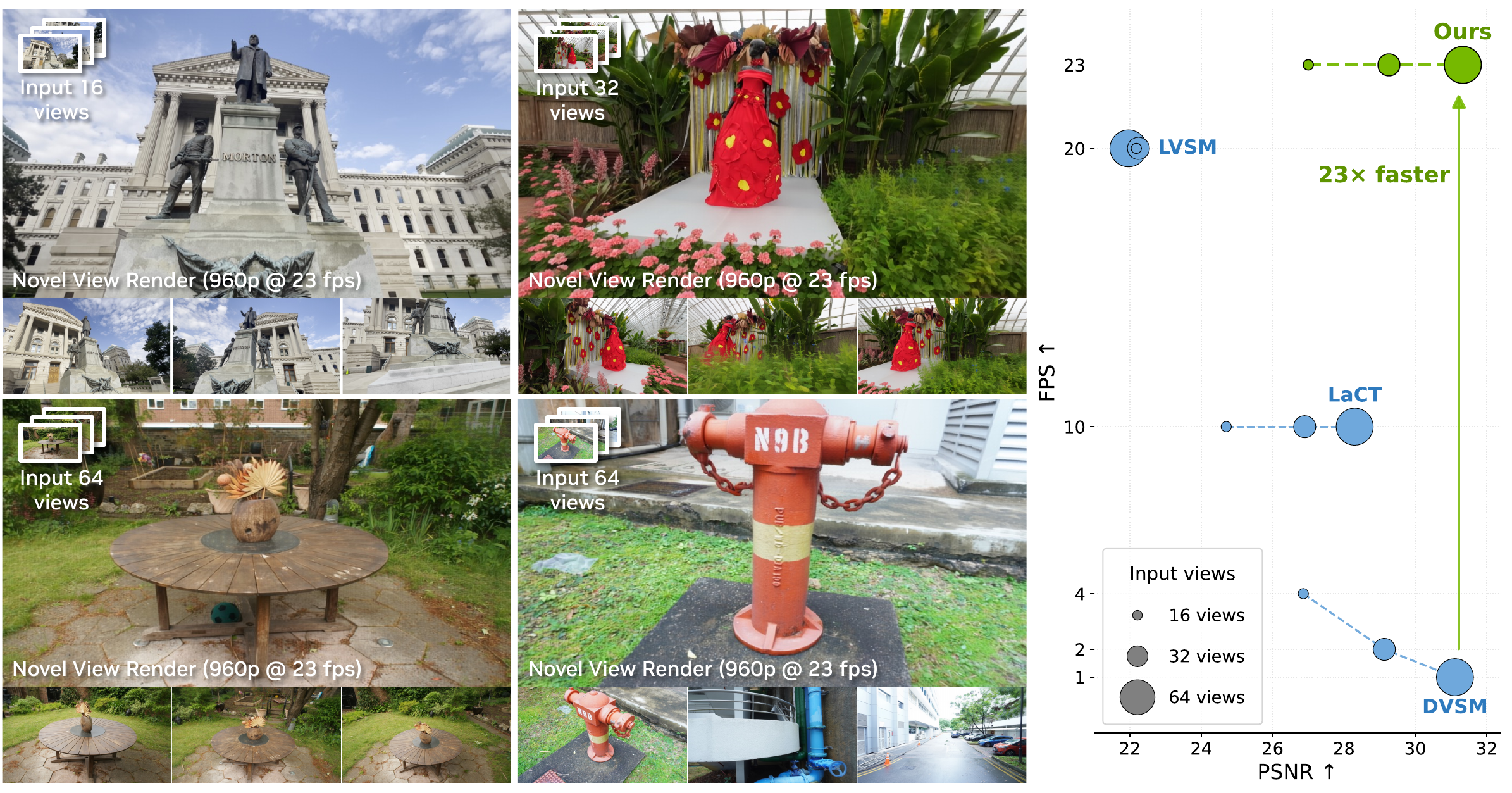}
    \vspace{-6mm}
    \caption{\textbf{T}est-\textbf{T}ime \textbf{B}aking (\textbf{TTB}) is a feed-forward, generalizable view synthesis model that bakes scene information from cross-view attention into a fast-weight MLP, achieving constant-complexity rendering with quality on par with decoder-only architectures. TTB sits on the Pareto frontier of PSNR vs.\ FPS, rendering up to $23\times$ faster than cross-attention baseline at comparable quality.}
    \label{fig:teaser}
\end{figure}

%% file: sec/0_abstract.tex
\begin{abstract}
\vspace{-2mm}
Decoder-only Large View Synthesis Models (LVSMs) with KV-cache have recently achieved state-of-the-art quality by regressing novel views from neural networks without reconstructing 3D geometry, while Test-Time Training (TTT) layers save computation but sacrifice quality.
We propose \textbf{T}est-\textbf{T}ime \textbf{B}aking (\textbf{TTB}), a strategy that learns to bake global scene information from cross-view attention into a lightweight MLP, enabling fast and high-fidelity novel-view decoding.
The first key is to repurpose the fast-weight MLP of TTT to learn the cross-view attention mapping explicitly from its input-output pairs, rather than replacing attention entirely, achieving the same rendering efficiency as TTT while delivering substantially higher quality.
The second key addresses a modality misalignment between context prefilling and novel-view rendering: we disentangle context tokens into camera-only context rendering tokens to serve as TTB input, and context ground-truth tokens with image content to serve as the TTB mapping target, ensuring the baked MLPs are relieved of the modality generalization challenge between context and novel-view processing.
Our method achieves up to \textbf{23x} rendering speedup with quality on par with the state-of-the-art across several benchmarks.
Project page: \url{https://test-time-baking.github.io}.
\end{abstract}

%% file: fig/motivation.tex
\begin{figure}[t]
    \centering
    \includegraphics[width=\linewidth]{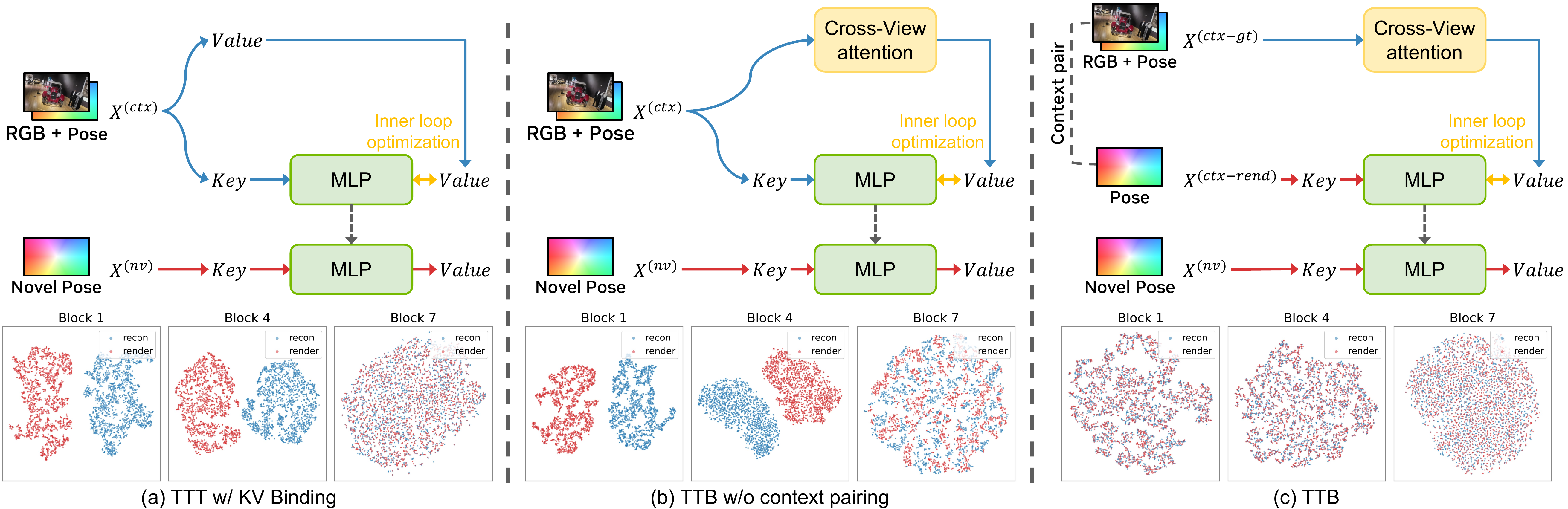}
    \vspace{-3mm}
    \caption{\textbf{Motivation.} We show three variants of the inner-loop optimization in TTT-style fast-weight MLPs. \textbf{(a)} TTT with KV binding~\cite{lact} learns from raw value tokens projected from $X^{(\text{ctx})}$. \textbf{(b)} TTB without context pairing targets the cross-view attention output instead, collaborating with attention rather than replacing it. \textbf{(c)} Our full TTB additionally pairs each context view: $X^{(\text{ctx-gt})}$ carries image content and supplies the cross-view attended target, while $X^{(\text{ctx-rend})}$ is camera-only and matches the modality of $X^{(\text{nv})}$. The t-SNE plots show that only (c) aligns the modalities from the first block.}
    \label{fig:motivation}
\end{figure}

%% file: sec/1_intro.tex
\section{Introduction}
\label{sec:intro}

Our research studies the task of novel-view synthesis conditioned on a set of context images with the powerful Large View Synthesis Models (LVSMs)~\cite{lvsm}.
LVSMs have recently emerged with several immediate follow-up improvements~\cite{efflvsm,svsm,clift,lact,dvsm}, achieving state-of-the-art novel-view synthesis quality across various input conditions~\cite{scenetok,rayzer,thely,truesn}.
Their geometry-free scene representations and Transformer-based renderers are fully learning-based, making them favorable for scaling with increasing amounts of training data.
The minimized 3D inductive bias also shows advantages when scenes are challenging for existing Transformer-compatible geometry-based representations (e.g., 3DGS~\cite{3dgs}), such as surfaces with strong mirror reflections or translucent material.

We are especially interested in the decoder-only variant of LVSM~\cite{lvsm} for its high-fidelity.
DVSM~\cite{dvsm} further introduces KV-cache to the cross-view attention layers, showing both quality and efficiency improvement.
The KV-cache serves as a global scene state for novel-view queries, but the rendering time is still impractical under dense context view conditions due to the linear complexity per token when attending to the KV-cache.
LaCT~\cite{lact} takes a different approach by replacing cross-view attention layers with Test-Time Training (TTT)~\cite{ttt} layers, as illustrated in Fig.~\ref{fig:motivation}a.
TTT performs inner-loop optimization to adapt fast-weight MLPs~\cite{fastweight} to directly map queries to their values without the quadratic-complexity attention, which can also be interpreted as a variant of linear attention~\cite{tttkvbind}.
Despite achieving constant-complexity rendering speed, TTT significantly sacrifices rendering quality.
A naive remedy for LaCT is to initialize from pretrained decoder-only model weights, which does improve quality, but the gap to the pretrained model remains significant.

We argue that this quality gap can be minimized by repurposing the fast-weight MLP inner-loop optimization as ``baking'' the information from cross-view attention, rather than replacing cross-view attention entirely as in TTT.
We dub our strategy Test-Time Baking (TTB), illustrated in Fig.~\ref{fig:motivation}b.
The key motivation is to trade context-view processing time for faster and higher-fidelity novel-view rendering, achieving the Pareto frontier of quality and speed shown in Fig.~\ref{fig:teaser}.
Specifically, the inner-loop learning target becomes the output tokens from cross-view attention, instead of the raw value tokens used in TTT.
This formulation also generalizes beyond the linear-attention interpretation of TTT~\cite{tttkvbind}.
In principle, the learning target of our TTB can also be tokens from arbitrarily non-linear layers or tokens enriched with additional scene information.

Furthermore, we identify a \emph{modality misalignment} issue inherent to the context prefilling of MLP fast-weight for novel-view synthesis.
In the reconstruction stage (context prefilling), the fast-weight MLPs are updated using context tokens with both camera and image information.
In the rendering stage, however, the updated MLPs take novel-view tokens coming from camera-only input without any image content.
This conventional pipeline poses a reconstruction-rendering input modality misalignment challenge.

Our solution to modality misalignment is simple yet effective.
As illustrated in Fig.~\ref{fig:motivation}c, we pair each context view into \textit{i)} context ground-truth tokens, derived from input cameras and images that produce the global scene information to be baked, and \textit{ii)} context rendering tokens, derived from camera-only input that simulates novel-view rendering conditions.
At each TTB layer, the fast-weight MLP is optimized to map context rendering tokens to their corresponding context ground-truth tokens, which carry image and cross-view attended information.
The optimized TTB-MLP is then shared between context rendering tokens and the downstream novel-view rendering tokens for retrieving scene information, ensuring that their conditioning remains identical across all layers. 

Interestingly, TTB draws a natural parallel with traditional implicit radiance and light field reconstruction, where MLPs are optimized to map spatial coordinates to radiance values.
In TTB, the fast-weight MLP instead is optimized to map ray-patch latent coordinates to their appearance latent representations, which effectively performs few-step light-field optimization in latent space.
This perspective suggests that TTB interpolates appearance latents for novel viewpoints from their ray-patch latent coordinates, motivating a convergence-oriented inner loop over one-shot summarization. In summary, our contributions are as follows:
\begin{itemize}
    \item We propose Test-Time Baking, which retargets the inner-loop optimization of the TTT fast-weight MLP from raw value tokens to cross-view attention outputs, so that the fast-weight MLP collaborates with attention rather than replacing it.
    \item We identify a modality misalignment between the reconstruction and rendering stages of MLP fast-weight prefilling, and resolve it via a context pair that disentangles each context view into a ground-truth and a camera-only rendering token.
    \item We draw a connection between TTB's inner loop and light field optimization in latent space, which motivates a convergence-oriented training recipe (multi-step MSE) that we find particularly effective for TTB.
\end{itemize}

%% file: sec/2_related.tex
\section{Related work}
\label{sec:related}

\textbf{Per-Scene optimization.}
Early approaches to novel-view synthesis perform per-scene optimization, fitting differentiable scene representations to observed images via differentiable rendering.
Coordinate-based MLPs~\cite{nerf,mipnerf,zipnerf}, voxels~\cite{dvgo,plenoxels,tensorf,instant-ngp,svraster,lin2024frugalnerf}, and the landmark 3DGS~\cite{3dgs} as the scene representation have been widely explored.
Despite impressive quality, per-scene optimization does not scale with the increasing training scenes and degrades quickly under fewer input views.


\textbf{Geometry-based feed-forward models.}
To overcome the scalability limitations and lengthy procedure of per-scene optimization, learning-based methods train networks to predict 3D representations from images.
Early works estimated multi-plane images via cost volumes~\cite{mvsnet,pmsnet,gwcnet,cvpmvsnet,casmvsnet,llff}, while NeRF-based generalizable methods~\cite{mvsnerf,pixelnerf,ibrnet,neuray} extended coverage to wide-angle viewpoints at limited quality.
The advent of 3DGS~\cite{3dgs} reframed the problem as pixel-aligned GS prediction~\cite{pixelsplat,depthsplat,mvsplat,flash3d,lgm,anysplat}, achieving promising results, and Large Reconstruction Models (LRMs)~\cite{lrm} further demonstrated Transformer scalability across various 3D representations~\cite{triplane,meshlrm,gslrm}.
One fundamental limitation of handcrafted 3D is their degradation on surfaces or materials they are not designed for, whereas the geometry-free representations we study can learn them given data.
Furthermore, explicit 3D geometry lacks a compact latent representation, and efforts to manage their linearly growing output size under more input views~\cite{longlrm,longlrmpp,tttlrm,tokengs} remain a symptom of this constraint.
Our TTB bakes scenes into constant-size MLPs at higher quality.

\textbf{Geometry-free feed-forward models.}
An alternative paradigm sidesteps the limitations of geometry-based models by representing scenes as latent representations and rendering via neural networks.
Early methods~\cite{attnrend,gpnr,lfnr,lfn,geofreevs,srt} were constrained by limited network capacity, while Large View Synthesis Models (LVSMs)~\cite{lvsm} broke this barrier by leveraging powerful Transformers.
Encoder-decoder and decoder-only LVSM variants are explored.
The encoder-decoder LVSM~\cite{lvsm} encodes context views into scene tokens and decodes novel views conditioned on the encoded scene, with subsequent works refining scene encoding strategies~\cite{efflvsm,svsm,clift} and extending to generative~\cite{scenetok} and uncalibrated camera~\cite{rayzer,erayzer,thely,truesn} settings.
The decoder-only LVSM~\cite{lvsm}, on the other hand, uses a single network for both context prefilling and novel-view rendering.
Despite higher fidelity, its original form inefficiently processes context views independently for each novel view.
DVSM~\cite{dvsm} addresses this by prefilling context views into a KV-cache shared across all novel views, achieving state-of-the-art quality under both sparse and dense view settings.
However, the linearly growing KV-cache size makes rendering speed impractical for larger numbers of input views.
LaCT~\cite{lact} replaces cross-view attention with Test-Time Training layers to achieve constant rendering complexity, but at a significant quality cost compared to DVSM.
Our Test-Time Baking achieves the same constant rendering complexity while substantially closing the quality gap left by LaCT.

\textbf{Test-time Training layers.}
TTT~\cite{ttt} has recently emerged as a popular alternative to attention for its linear time and constant space complexity.
We refer interested readers to \cite{tttkvbind} for an in-depth review and analysis.
We are particularly interested in the variant presented in LaCT~\cite{lact}, which has demonstrated success in accelerating various vision and 3D tasks~\cite{ttt3r,tttlrm,vittt,vggttt,zipmap}, albeit at the cost of quality relative to their attention-based counterparts.
Our TTB departs from this paradigm in two ways: rather than replacing attention, TTB learns directly from attention outputs; and rather than performing mere token-to-token information exchange, TTB explicitly learns to map camera-ray tokens to their corresponding image tokens, yielding improved novel-view synthesis quality.


%% file: fig/pipeline.tex
\begin{figure}[t]
    \centering
    \includegraphics[width=\linewidth]{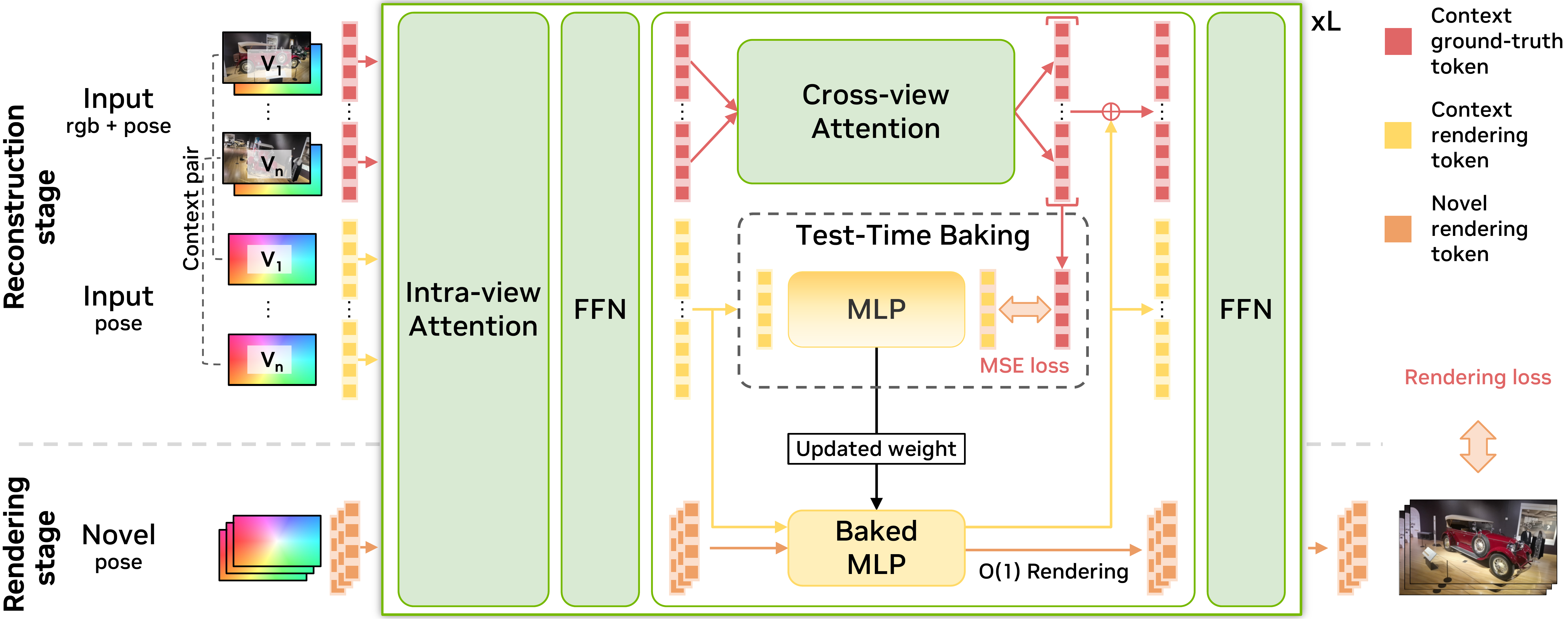}
    \vspace{-3mm}
    \caption{\textbf{Pipeline.} Our model stacks $L$ identical blocks of Intra-view Attention, FFN, Test-Time Baking, and FFN. In the reconstruction stage, each context view is duplicated into a context pair, where the context ground-truth token $X^{(\text{ctx-gt})}$ carries RGB and pose while the context rendering token $X^{(\text{ctx-rend})}$ is pose-only and matches the modality of the novel rendering token $X^{(\text{nv})}$. At each TTB layer, the inner-loop \emph{update} bakes the fast-weight MLP via MSE loss to map $X^{(\text{ctx-rend})}$ toward the cross-view attention output of $X^{(\text{ctx-gt})}$. 
    The baked MLP is \emph{applied} to serve three token streams: it propagates both $X^{(\text{ctx-rend})}$ and $X^{(\text{nv})}$ to the next layer, and provides feedback to $X^{(\text{ctx-gt})}$ as an auxiliary signal. In the novel-view rendering stage, each token retrieves global scene information in $O(1)$ from the baked MLP, instead of $O(N)$ via cross-attention. The rendering loss is backpropagated through the inner-loop baking process to train the model end-to-end.
    }
    \label{fig:pipeline}
    \vspace{-2mm}
\end{figure}

%% file: sec/3_0_approach.tex
\section{Method}
\label{sec:approach}

We start by formally introducing our task and the general pipeline in Sec.~\ref{ssec:overview}.
We detail the decoder-only architecture we build on in Sec.~\ref{ssec:dvsm}.
Finally, we present Test-Time Baking in Sec.~\ref{ssec:ttb}.

%% file: sec/3_1_overview.tex
\subsection{Overview}
\label{ssec:overview}

Given a set of $V$ context images with known poses and intrinsics $\{\Image_i\in\Real^{H\times W\times 3}, \Pose_i\in \mathbb{SE}(3), \Intr_i\in \Real^{3\times 3}\}_{i=1}^V$, our goal is to synthesize novel views for an arbitrary camera query $(\Pose', \Intr')$.
We focus on reconstruction rather than generation, assuming novel views can be directly regressed from the context views.
Approaches to this end generally decompose into two stages:
\begin{equation}
    S = \recon(\Image, \Pose, \Intr),  \quad 
    \bar{\Image} = \rend(\Pose', \Intr', S),
\end{equation}
where $\recon(\cdot)$ processes context views into a method-specific scene representation $S$, and $\rend(\cdot)$ synthesizes the novel view $\bar{\Image}$ conditioned on $S$.
Recent feed-forward models instantiate $\recon(\cdot)$ as a Transformer.
The Large Reconstruction Model (LRM)~\cite{lrm} family employs a human-defined 3D and renderer, while the Large View Synthesis Model (LVSM)~\cite{lvsm} series defines $S$ as a latent state and implements $\rend(\cdot)$ as a neural network.
As both $\recon(\cdot)$ and $\rend(\cdot)$ are differentiable, models are trained end-to-end against ground-truth novel views $\Image'$ by:
\begin{equation} \label{eq:objective}
    \mathcal{J} = \text{MSE}(\bar{\Image}, \Image') + \lambda \cdot \text{Percep}(\bar{\Image}, \Image')\,,
\end{equation}
where $\text{Percep}$ is the neural perceptual loss~\cite{perceptual} and $\lambda$ balances the photometric objectives.

%% file: sec/3_2_DVSM.tex
\subsection{Decoder-only View Synthesis Model}
\label{ssec:dvsm}

Our method builds on the Decoder-only View Synthesis Model (DVSM)~\cite{dvsm}, which uses a shared Transformer decoder for both $\recon(\cdot)$ and $\rend(\cdot)$, and represents the scene $S$ as KV caches.
Specifically, the model consists of $L$ repeated blocks, each comprising a sequence of layers:
\begin{equation} \label{eq:dvsm}
    \token_\ell = (\ffn_{\ell,2} \circ \attncross_\ell \circ \ffn_{\ell,1} \circ \attnintra_\ell)(\token_{\ell-1})\,,
\end{equation}
where $\attnintra(\cdot)$ and $\attncross(\cdot)$ perform intra-view and cross-view attention, respectively, and $\ffn(\cdot)$ denotes a fully connected layer.
Residual connections~\cite{resnet} and QK-normalization~\cite{qknorm} are applied for training stability.
During reconstruction, context-view tokens are prefilled into a KV-cache at each $\attncross(\cdot)$.
During rendering, novel-view tokens query the cached KV to retrieve global scene information from each $\attncross(\cdot)$ layer.

Regarding input tokenization, camera intrinsics and poses are first converted into Pl\"ucker ray maps~\cite{plucker} for linear patch embedding. Context-view and novel-view tokens are initialized as:
\begin{equation} \label{eq:inp_tokens}
    \tokenctx_0 = \text{LayerNorm}(\PEray(\Pose, \Intr) + \PErgb(\Image)),  \quad 
    \tokennv_0 = \text{LayerNorm}(\PEray(\Pose', \Intr'))\,,
\end{equation}
where $\PEray$ and $\PErgb$ are linear patch projections for the Pl\"ucker ray maps and context images, respectively, Here, $\tokenctx_0\in\Real^{V\times\frac{H}{p}\times\frac{W}{p}\times D}$ denotes tokens from $V$ context views with patch size $p$ used for KV-cache prefilling, and $\tokennv_0\in\Real^{U\times\frac{H}{p}\times\frac{W}{p}\times D}$ denotes tokens for $U$ novel-view queries, each processed independently.
Finally, a shallow model head regresses pixel values as:
\begin{equation} \label{eq:rgb_head}
    \bar{\Image} = (\text{PixShuf} \circ \text{Sigmoid} \circ \text{Linear} \circ \text{LayerNorm})(\tokennv_L)\,,
\end{equation}
where $\text{PixShuf}(\cdot)$ denotes pixel shuffling~\cite{pixshufl} and $\tokennv_L$ is the last-layer tokens out of Eq.\ref{eq:dvsm}.

%% file: sec/3_3_0_TTB.tex
\subsection{Test-time Baking}
\label{ssec:ttb}

We present our technical contribution to improve the rendering-time efficiency of DVSM~\cite{dvsm} while retaining its high rendering quality.
The issue of DVSM is that its KV-cache size grows linearly with the number of context tokens, making it slow and impractical for novel-view queries under high-resolution or dense input context views.
A naive strategy is to replace each $\attncross(\cdot)$ in Eq.~\ref{eq:dvsm} with a Test-Time Training (TTT)~\cite{ttt} layer, achieving constant space and linear time complexity as demonstrated in LaCT~\cite{lact}.
However, we observe significant quality degradation even when finetuning the TTT variant from DVSM pretrained weights.
This motivates us to explore a better solution that improves upon TTT.
A full model pipeline is illustrated in Fig.~\ref{fig:pipeline}.

%% file: sec/3_3_1_repurpose.tex
\subsubsection{Repurposing TTT: collaborating instead of replacing attention}
\label{sssec:repurpose}

Our motivation is to trade reconstruction (prefilling) computation, which only needs to be performed once per scene, for improved rendering-time speed and quality.
The resulting solution is simple yet effective.
Rather than entirely replacing cross-view attention with TTT fast-weight MLP as in the LaCT~\cite{lact}, we repurpose TTT to explicitly learn from the input-output pairs out of cross-view attention.
Consequently, we no longer need to cache the KV; only the updated MLP fast weights are stored.
The space and time complexity of novel-view rendering is therefore identical to that of TTT.
We call our strategy Test-Time Baking (TTB), which can be viewed as a generalized form of TTT.

Formally, the fast-weight update rules of TTT and TTB are:
\begin{subequations} \label{eq:ttt_ttb_upd}
\begin{align}
    \theta^{\text{(TTT)}} &= \theta - \eta\nabla_\theta \mathcal{L}(f_{\theta}(\mathbf{W}_k\mathbf{X}), \mathcolor{downred}{\mathbf{W}_v\mathbf{X}}), \label{eq:ttt_upd}
    \\
    \theta^{\text{(TTB)}} &= \theta - \eta\nabla_\theta \mathcal{L}(f_{\theta}(\mathbf{W}_k\mathbf{X}), \mathcolor{upgreen}{g(\mathbf{X^*})}), \label{eq:ttb_upd}
\end{align}
\end{subequations}
respectively, where $\theta$ is the fast weights of the MLP $f(\cdot)$, $\eta$ and $\mathcal{L}$ are the inner-loop learning rate and objective, $\mathbf{W}_k$ and $\mathbf{W}_v$ are linear projection matrices, and $\mathbf{X}$ is the input tokens.
The difference between TTT and TTB lies in the mapping target: $\mathcolor{downred}{\mathbf{W}_v\mathbf{X}}$ and $\mathcolor{upgreen}{g(\mathbf{X^*})}$.
In principle, we can employ an arbitrary function $g$ or tokens $\mathbf{X^*}$ with additional modality for the mapping target $\mathcolor{upgreen}{g(\mathbf{X^*})}$.
Note that TTB falls back to TTT when $\mathbf{X^*}{=}\mathbf{X}$ and $g$ is a linear projection.
In our case, we instantiate $g(\cdot)$ as the $\attncross(\cdot)$ from DVSM, which shifts TTB beyond the linear-attention interpretation~\cite{tttkvbind} of TTT. 
Although this incurs additional reconstruction-time cost, once prefilled, rendering remains identical to TTT with substantially higher quality thanks to the enriched global information from $g(\cdot)$.


\begin{wrapfigure}{r}{0.255\textwidth}
    \vspace{-1.2em}
    \centering
    \includegraphics[width=0.255\textwidth]{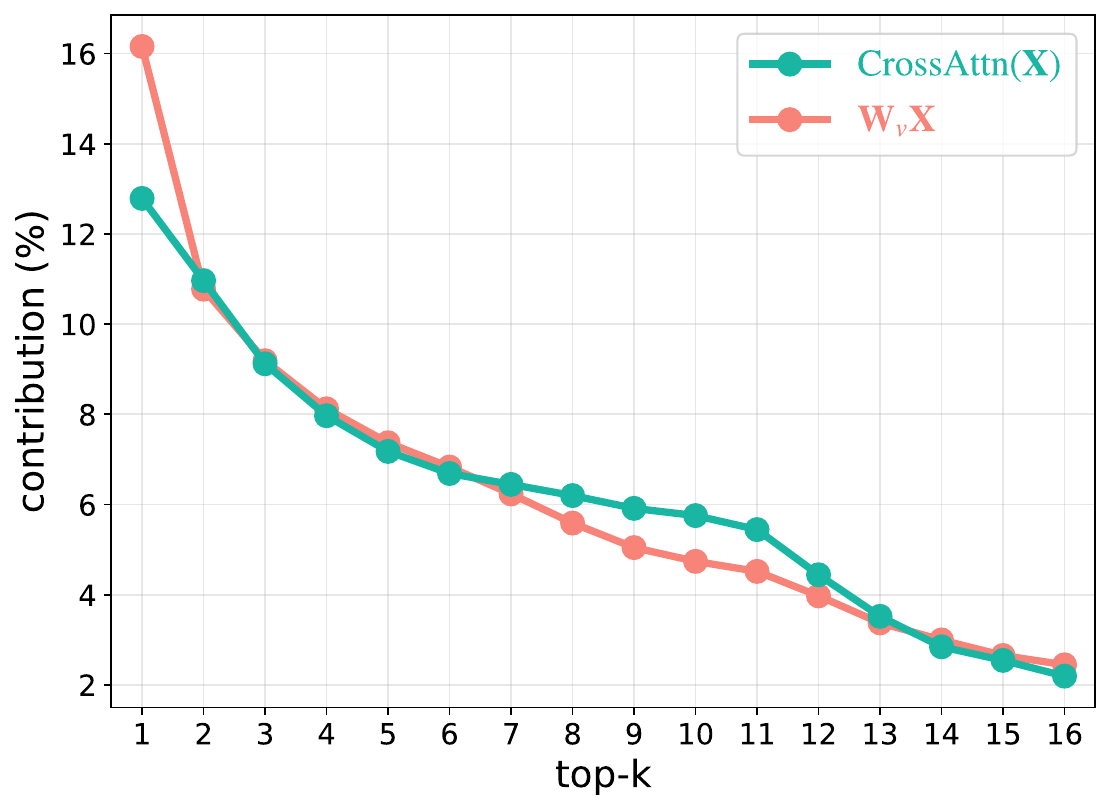}
    \vspace{-1.6em}
    \caption{\textbf{Per-rank context contribution.}}
    \label{fig:per_rank}
    \vspace{-1.4em}
\end{wrapfigure}
\textbf{More analysis.} We empirically validate that swapping the inner-loop target from $\mathcolor{downred}{\mathbf{W}_v\mathbf{X}}$ to $\mathcolor{upgreen}{\attncross(\mathbf{X})}$ broadens the \emph{effective view field} of the baked fast-weight MLP.
At TTB block~7 with $V{=}16$, we measure how much each context image contributes to the TTB output at the novel-view rendering tokens via input-gradient saliency, sorted and averaged across output positions (Fig.~\ref{fig:per_rank}).
The $\mathbf{W}_v\mathbf{X}$ variant peaks sharply at top-$1$ then decays, while $\attncross(\mathbf{X})$ flattens the peak and sustains a heavier mid-rank tail.
This matches our design intent: $\attncross(\mathbf{X})$ already aggregates across all $V$ context views, so each memorized pair inherits a wider receptive field.

%% file: sec/3_3_2_ctx_pair.tex
\subsubsection{Modality misalignment and context pair}
\label{sssec:ctx_pair}


The baked MLPs via Eq.~\ref{eq:ttb_upd} receive inputs of different modalities between the reconstruction and rendering stages.
As defined in Eq.~\ref{eq:inp_tokens}, context tokens $\tokenctx_0$ used for prefilling contain image content, whereas novel-view tokens $\tokennv_0$ are with camera-only information.
This \emph{modality misalignment} poses an unnecessary generalization challenge to the baked MLP.

We address this by disentangling input context tokens, $\tokenctx_0$ from Eq.\ref{eq:inp_tokens}, into a paired representation:

\begin{equation} \label{eq:ctx_pair}
    \tokenctxgt_0 = \text{LayerNorm}(\PEray(\Pose, \Intr) + \PErgb(\Image)), \;
    \tokenctxrend_0 = \text{LayerNorm}(\PEray(\Pose, \Intr)),
\end{equation}

where $\tokenctxrend_0$ are the \emph{context rendering tokens} with camera-only input, and $\tokenctxgt_0$ are the \emph{context ground-truth tokens} with additional RGB input.
We use $\tokenctxgt$ to serve TTB mapping target while using $\tokenctxrend$ to serve TTB input.
Specifically, at each TTB layer, the fast-weight MLP is baked to learn the mapping $\tokenctxrend \mapsto \attncross(\tokenctxgt)$ via the update rule in Eq.~\ref{eq:ttb_upd}.
The baked MLP is then shared across both context rendering $\tokenctxrend$ and the downstream novel-view rendering $\tokennv$ tokens to produce feature for the subsequent layers.
As visualized in Fig.~\ref{fig:motivation}, our context pairing ensures distribution alignment from the early layers, which facilitates feature interpolation.
Conversely, the distribution only aligned in the later layers without context pairing due to the modality misalignment between reconstruction and rendering stages.
Despite doubling the number of tokens during reconstruction, rendering-stage efficiency remains the same, which aligns with our motivation to trade reconstruction time for rendering-time quality.

%% file: sec/3_3_3_arch_details.tex
\subsubsection{TTB layer details}
\label{sssec:ttb_arch_details}

The design of our TTB layer is inspired by the TTT layer in LaCT~\cite{lact}.
We remove the asymmetric QK projection and instead use a shared linear projection to produce queries and keys.
We adopt SwiGLU-MLP~\cite{glu} as the fast-weight MLP architecture, and apply an additional linear layer to predict per-token learning rates.
For inner-loop optimization, we employ Muon~\cite{muon} optimizer with weight normalization~\cite{weightnormalization} for a predefined $T$ steps.
We find that using MSE as the inner-loop objective for our TTB yields better results than the unnormalized dot-product loss used in LaCT~\cite{lact}'s TTT.
We also find that learning the residual to the mapping target is easier to bake than learning the target directly.
Specifically, the final TTB update rule is:

\begin{equation} \label{eq:final_ttb}
\theta_{t+1} = \theta_t - \mathrm{Muon}\left( \nabla_{\theta_t} \left\| f_{\theta_t}\left(\mathbf{W}_k\tokenctxrend\right) - \mathbf{W}_v \left(\attncross(\tokenctxgt) - \tokenctxrend\right) \right\|_2 \right) ~,
\end{equation}

where $\mathrm{Muon}(\cdot)$ performs Muon gradient orthogonalization~\cite{muon}, $\mathbf{W}_k$ and $\mathbf{W}_v$ are linear projection matrices for the TTB input and mapping target.
The final baked weights $\theta_T$ are then applied to update both the context rendering and novel-view rendering tokens and provide feedback to the context ground-truth tokens (the output arrows from the baked MLP in Fig.~\ref{fig:pipeline}).

%% file: sec/3_3_4_infer_strategy.tex
\subsubsection{Fast-weight local-duplicate inference}
\label{sssec:infer_strategy}
The baked fast weights have a fixed capacity, causing quality to plateau as the number of context
views $V$ grows. We address this at inference time by clustering the context views into $K$
spatially-local groups via k-means on per-view camera poses, with $K=V/B$ for a nominal budget of
$B$ views per group. Each group is reconstructed independently by the unmodified model, so that
cross-view attention and the inner-loop update only involve the views of that group, yielding
per-group, per-layer fast weights $\{\{\theta_{k,\ell}\}_{\ell=1}^{L}\}_{k=1}^{K}$. At rendering
time, each novel view is routed to the group with the nearest centroid and queries only the
corresponding copy. Rendering thus remains $O(1)$ per view and each copy only has to absorb about
$B$ views, while the stored state grows linearly with $K$.

%% file: fig/qual_dl3dv.tex

\begin{figure}[!t]
    \centering
    \includegraphics[width=\linewidth]{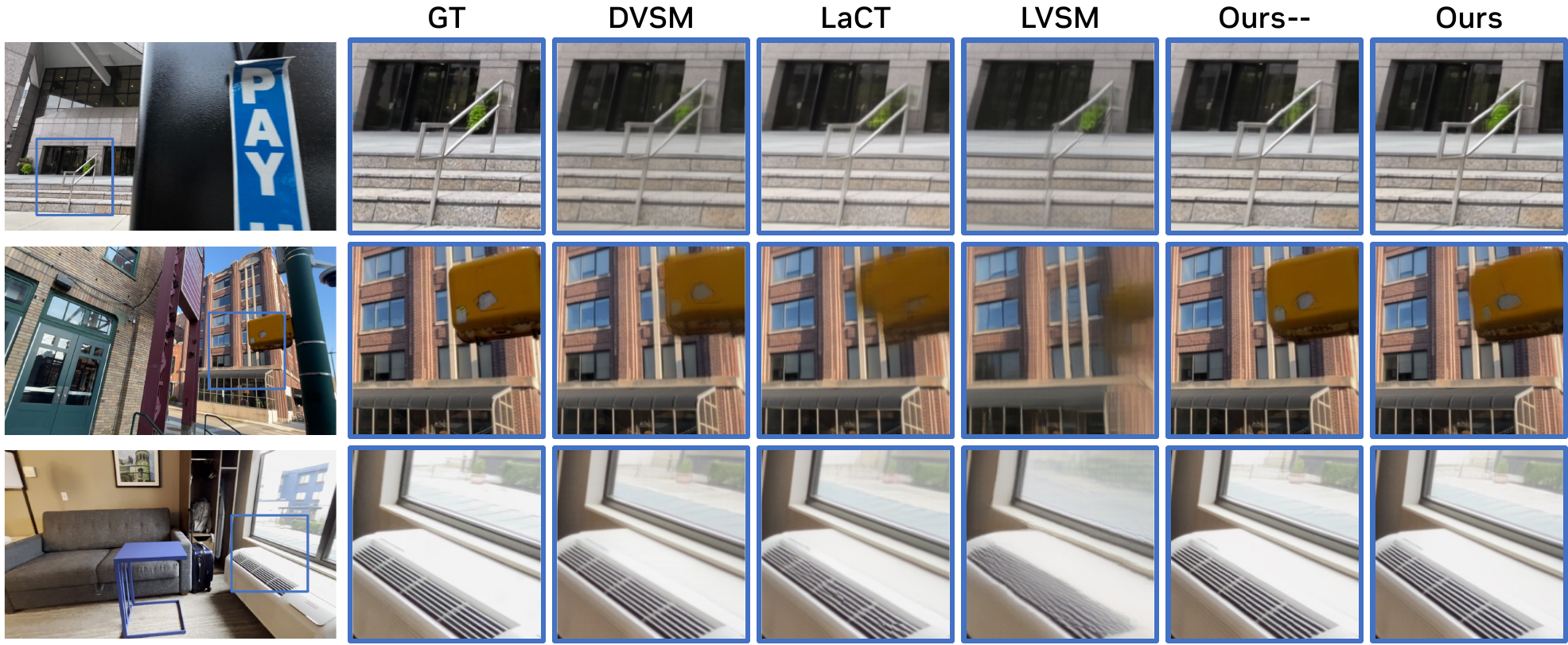}
    \vspace{-3mm}
    \caption{\textbf{Qualitative comparison on DL3DV.} TTB preserves fine details on par with DVSM~\cite{dvsm}, while LaCT~\cite{lact} and LVSM~\cite{lvsm} produce visible blur.}
    \label{fig:qual_dl3dv}
\end{figure}

%% file: tab/re10k_dl3dv.tex
\begin{table}[t]
\centering
\vspace{-2mm}
\caption{{\bf Quantitative comparison on DL3DV.}}
\label{tab:dl3dv}
\setlength{\tabcolsep}{3pt}
\begin{adjustbox}{max width=\linewidth}
  \begin{tabular}{ll ccccc ccccc ccccc}
  \toprule
  & &
  \multicolumn{5}{c}{\bf 16 input views} &
  \multicolumn{5}{c}{\bf 32 input views} &
  \multicolumn{5}{c}{\bf 64 input views} \\
  \cmidrule(lr){3-7}
  \cmidrule(lr){8-12}
  \cmidrule(lr){13-17}
  Method & Venue &
  PSNR$\uparrow$ & SSIM$\uparrow$ & LPIPS$\downarrow$ & Time$\downarrow$ & FPS$\uparrow$ &
  PSNR$\uparrow$ & SSIM$\uparrow$ & LPIPS$\downarrow$ & Time$\downarrow$ & FPS$\uparrow$ &
  PSNR$\uparrow$ & SSIM$\uparrow$ & LPIPS$\downarrow$ & Time$\downarrow$ & FPS$\uparrow$ \\
  \midrule
    3DGS~\cite{3dgs} & ToG'23 &
    21.20 & 0.708 & 0.264 & >600s & \cellcolor{gray!25}>50 &
    23.60 & 0.779 & 0.213 & >600s & \cellcolor{gray!25}>50 &
    26.43 & 0.854 & 0.167 & >600s & \cellcolor{gray!25}>50 \\
    \midrule
    LongLRM~\cite{longlrm} & ICCV'25 &
    22.66 & 0.740 & 0.292 & \cellcolor{gray!25}0.4s & \cellcolor{gray!25}>50 &
    24.10 & 0.783 & 0.254 & \cellcolor{gray!25}1s & \cellcolor{gray!25}>50 &
    24.77 & 0.804 & 0.239 & \cellcolor{gray!25}3s & \cellcolor{gray!25}>50 \\
    LongLRM++~\cite{longlrmpp} & arxiv'25 &
    24.40 & 0.795 & 0.231 & \cellcolor{gray!25}1.6s & 14 &
    26.43 & 0.846 & 0.180 & \cellcolor{gray!25}4.7s & 14 &
    27.30 & 0.869 & 0.161 & \cellcolor{gray!25}16s & 14 \\
    tttLRM~\cite{tttlrm} & CVPR'26 &
    23.60 & 0.784 & 0.255 & 3.6s & - &
    25.07 & 0.822 & 0.215 & 7.2s & - &
    25.95 & 0.844 & 0.195 & 15s & - \\
    \midrule
    LVSM~\cite{lvsm} & ICLR'25 &
    22.18 & 0.694 & 0.351 & \cellcolor{yellow!25}3.8s & \cellcolor{yellow!25}20 &
    22.24 & 0.691 & 0.352 & \cellcolor{yellow!25}15s & 20 &
    21.96 & 0.678 & 0.364 & \cellcolor{yellow!25}59s & 20 \\
    LaCT~\cite{lact} & ICLR'26 &
    24.70 & 0.793 & 0.224 & \cellcolor{red!25}1.9s & 10 &
    26.90 & 0.837 & 0.185 & \cellcolor{red!25}3.7s & 10 &
    28.30 & 0.857 & 0.169 & \cellcolor{orange!25}7.3s & 10 \\
    DVSM~\cite{dvsm} & {\footnotesize arXiv'26} &
    \cellcolor{orange!25}26.86 & \cellcolor{red!25}0.852 & \cellcolor{red!25}0.170 & 4.4s & 4 &
    \cellcolor{orange!25}29.13 & \cellcolor{red!25}0.893 & \cellcolor{red!25}0.129 & \cellcolor{yellow!25}15s & 2 &
    \cellcolor{orange!25}31.11 & \cellcolor{red!25}0.920 & \cellcolor{red!25}0.104 & 60s & 1 \\
    \midrule
    Ours{-}{-} &  &
    \cellcolor{yellow!25}26.26 & \cellcolor{yellow!25}0.833 & \cellcolor{yellow!25}0.191 & \cellcolor{red!25}1.9s & \cellcolor{red!25}23 &
    28.12 & 0.867 & 0.158 & \cellcolor{red!25}3.7s & \cellcolor{red!25}23 &
    29.29 & 0.882 & 0.144 & \cellcolor{orange!25}7.5s & \cellcolor{red!25}23 \\
    Ours &  &
    \cellcolor{red!25}27.00 & \cellcolor{orange!25}0.849 & \cellcolor{orange!25}0.177 & 5.4s & \cellcolor{red!25}23 &
    \cellcolor{yellow!25}28.92 & \cellcolor{yellow!25}0.882 & \cellcolor{yellow!25}0.145 & 18s & \cellcolor{red!25}23 &
    \cellcolor{yellow!25}30.20 & \cellcolor{yellow!25}0.897 & \cellcolor{yellow!25}0.131 & 65s & \cellcolor{red!25}23 \\
    Ours (duplicate) &  &
    - & - & - & - & - &
    \cellcolor{red!25}29.26 & \cellcolor{orange!25}0.890 & \cellcolor{orange!25}0.136 & 18s & \cellcolor{red!25}23 &
    \cellcolor{red!25}31.33 & \cellcolor{orange!25}0.919 & \cellcolor{orange!25}0.108 & 65s & \cellcolor{red!25}23 \\
  \bottomrule
  \end{tabular}
\end{adjustbox}
\end{table}

\begin{table}[!t]
\centering
\vspace{-2mm}
\caption{{\bf Two input views on Re10k at 512$\times$512.} Results are denoted as PSNR$\uparrow$ / SSIM$\uparrow$ / LPIPS$\downarrow$.}
\label{tab:re10k}
\begin{adjustbox}{max width=\linewidth}
  \begin{tabular}{
    @{}
    cccccc
    @{}}
    \toprule
    LVSM~\cite{lvsm} & Decoder-only LVSM~\cite{lvsm} & Efficient-LVSM~\cite{efflvsm} & DVSM~\cite{dvsm} & Ours{-}{-} & Ours \\
    \midrule
    28.55 / 0.894 / 0.173 & 
    29.53 / 0.904 / 0.141 & 
    29.86 / 0.905 / 0.147 & 
    31.17 / 0.924 / 0.129 & 
    31.00 / 0.921 / 0.130 & 
    {\bf 31.28 / 0.926 / 0.127} \\
    \bottomrule
  \end{tabular}
\end{adjustbox}
\vspace{-3mm}
\end{table}

%% file: fig/qual_wild.tex
\begin{figure}[t]
    \centering
    \includegraphics[width=\linewidth]{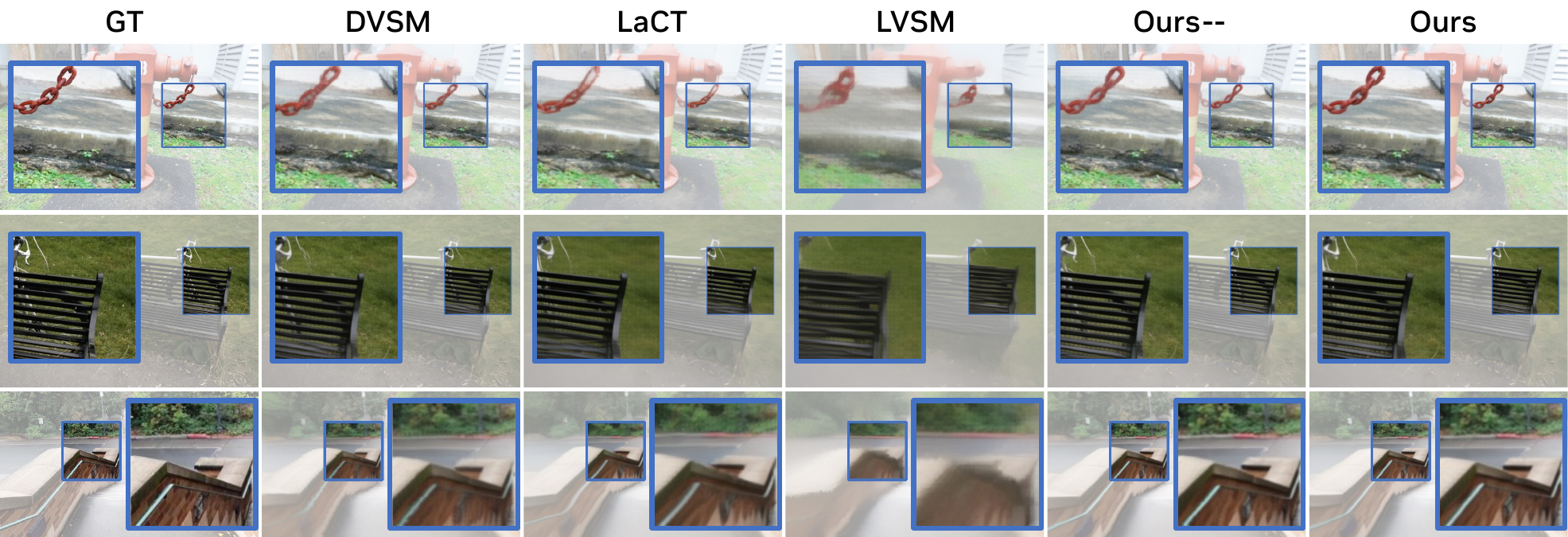}
    \vspace{-5mm}
    \caption{\textbf{Qualitative comparison on in-the-wild datasets.} TTB generalizes to out-of-domain scenes and preserves fine details consistent with the in-domain results.}
    \label{fig:qual_wild}
\end{figure}

%% file: tab/in_the_wild.tex
\begin{table}[t]
\centering
\caption{{\bf Quantitative comparison on in-the-wild datasets.}}
\label{tab:in_the_wild}
\begin{adjustbox}{max width=\linewidth}
\begin{tabular}{
    l@{\hskip 3pt}c
    ccc c 
    ccc c 
    ccc c  cc
}
    \toprule
    &&
    \multicolumn{3}{c}{\bf MipNerf360~\cite{mip-nerf360}} &&
    \multicolumn{3}{c}{\bf Free~\cite{f2nerf}} &&
    \multicolumn{3}{c}{\bf Hike~\cite{hike}} &&
    \multicolumn{2}{c}{\bf Efficiency} \\
    \cmidrule(lr){3-5}\cmidrule(lr){7-9}\cmidrule(lr){11-13}\cmidrule(lr){15-16}
    Method & Size &
    PSNR$\uparrow$&SSIM$\uparrow$&LPIPS$\downarrow$ &&
    PSNR$\uparrow$&SSIM$\uparrow$&LPIPS$\downarrow$ &&
    PSNR$\uparrow$&SSIM$\uparrow$&LPIPS$\downarrow$ &&
    Time$\downarrow$ & FPS$\uparrow$\\
    \midrule
    \multicolumn{14}{@{}l@{}}{\bf \underline{Input: 64 views $\times$ 518{,}400 pixels per view (960p)}} \\
    LVSM & {\scriptsize 165M} &
    20.07 & 0.459 & 0.559 &&
    20.27 & 0.517 & 0.444 &&
    18.60 & 0.433 & 0.502 &&
    \cellcolor{yellow!25}59s & \cellcolor{yellow!25}20 \\
    LaCT & {\scriptsize 379M} &
    23.01 & 0.557 & 0.435 &&
    24.61 & 0.680 & 0.265 &&
    22.39 & 0.648 & 0.305 &&
    \cellcolor{red!25}7.3s & 10 \\
    DVSM & {\scriptsize 163M} &
    \cellcolor{orange!25}24.76 & \cellcolor{orange!25}0.660 & \cellcolor{red!25}0.337 &&
    \cellcolor{yellow!25}25.57 & \cellcolor{orange!25}0.756 & \cellcolor{orange!25}0.205 &&
    \cellcolor{yellow!25}23.10 & \cellcolor{yellow!25}0.697 & \cellcolor{yellow!25}0.262 &&
    60s & 1 \\
    Ours{-}{-} & {\scriptsize 244M} &
    23.93 & 0.597 & 0.397 &&
    25.28 & 0.708 & 0.253 &&
    22.53 & 0.660 & 0.300 &&
    \cellcolor{orange!25}7.5s & \cellcolor{red!25}23 \\
    Ours & {\scriptsize 271M} &
    \cellcolor{yellow!25}24.49 & \cellcolor{yellow!25}0.624 & \cellcolor{yellow!25}0.375 &&
    \cellcolor{orange!25}26.05 & \cellcolor{yellow!25}0.744 & \cellcolor{yellow!25}0.225 &&
    \cellcolor{orange!25}23.77 & \cellcolor{orange!25}0.707 & \cellcolor{orange!25}0.259 &&
    65s & \cellcolor{red!25}23 \\
    Ours (duplicate) & {\scriptsize 271M} &
    \cellcolor{red!25}25.05 & \cellcolor{red!25}0.667 & \cellcolor{orange!25}0.341 &&
    \cellcolor{red!25}26.82 & \cellcolor{red!25}0.800 & \cellcolor{red!25}0.187 &&
    \cellcolor{red!25}25.37 & \cellcolor{red!25}0.782 & \cellcolor{red!25}0.195 &&
    65s & \cellcolor{red!25}23 \\
    \midrule
    \multicolumn{14}{@{}l@{}}{\bf \underline{Input: 64 views $\times$ 512 pixels at image longer side}} \\
    LagerNVS & {\scriptsize 1052M} &
    20.98 & 0.481 & 0.503 &&
    19.50 & 0.486 & 0.396 &&
    18.54 & 0.436 & 0.415 &&
    \cellcolor{orange!25}2.3s & 7 \\
    DVSM & {\scriptsize 163M} &
    \cellcolor{red!25}24.04 & \cellcolor{red!25}0.607 & \cellcolor{red!25}0.422 &&
    \cellcolor{yellow!25}24.72 & \cellcolor{orange!25}0.717 & \cellcolor{orange!25}0.268 &&
    \cellcolor{yellow!25}23.39 & \cellcolor{orange!25}0.705 & \cellcolor{yellow!25}0.264 &&
    \cellcolor{yellow!25}5.0s & \cellcolor{yellow!25}12 \\
    Ours{-}{-} & {\scriptsize 244M} &
    23.18 & 0.552 & 0.478 &&
    24.27 & 0.665 & 0.323 &&
    22.41 & 0.646 & 0.295 &&
    \cellcolor{red!25}1.8s & \cellcolor{red!25}67 \\
    Ours & {\scriptsize 271M} &
    \cellcolor{yellow!25}23.62 & \cellcolor{yellow!25}0.568 & \cellcolor{yellow!25}0.463 &&
    \cellcolor{orange!25}24.92 & \cellcolor{yellow!25}0.692 & \cellcolor{yellow!25}0.299 &&
    \cellcolor{orange!25}23.52 & \cellcolor{yellow!25}0.687 & \cellcolor{orange!25}0.262 &&
    6.3s & \cellcolor{red!25}67 \\
    Ours (duplicate) & {\scriptsize 271M} &
    \cellcolor{orange!25}24.03 & \cellcolor{orange!25}0.602 & \cellcolor{orange!25}0.434 &&
    \cellcolor{red!25}25.45 & \cellcolor{red!25}0.733 & \cellcolor{red!25}0.267 &&
    \cellcolor{red!25}24.96 & \cellcolor{red!25}0.760 & \cellcolor{red!25}0.206 &&
    6.3s & \cellcolor{red!25}67 \\
    \bottomrule
\end{tabular}
\end{adjustbox}
\vspace{-3mm}
\end{table}

%% file: tab/ablation_ttb.tex
\begin{table}[t]
\centering
\small
\setlength{\tabcolsep}{4pt}
\renewcommand{\arraystretch}{1.05}
\setlength{\aboverulesep}{1pt}
\setlength{\belowrulesep}{1pt}
\caption{\textbf{Ablation study from TTT to TTB on DL3DV.} ({\setlength{\fboxsep}{1pt}\colorbox{green!60!black!15}{\,light\,}}=worst, {\setlength{\fboxsep}{1pt}\colorbox{green!60!black!55}{\,dark\,}}=best).}
\label{tab:ablation_ttb}
\resizebox{\linewidth}{!}{%
\begin{tabular}{lccc|ccc|ccc}
\toprule
\multirow{2}{*}{Method} & \multicolumn{3}{c|}{Quality ($V{=}32$)} & \multicolumn{3}{c|}{Recon-time$\downarrow$} & \multicolumn{3}{c}{Rend-fps$\uparrow$} \\
\cmidrule(lr){2-4} \cmidrule(lr){5-7} \cmidrule(lr){8-10}
& PSNR$\uparrow$ & SSIM$\uparrow$ & LPIPS$\downarrow$ & $V{=}16$ & $V{=}32$ & $V{=}64$ & $V{=}16$ & $V{=}32$ & $V{=}64$ \\
\midrule
(a) = DVSM~\cite{dvsm}
 & \cellcolor{green!60!black!51}24.68 & \cellcolor{green!60!black!55}0.776 & \cellcolor{green!60!black!55}0.220
 & \cellcolor{green!60!black!33}1.300 & \cellcolor{green!60!black!25}4.907 & \cellcolor{green!60!black!20}18.396
 & \cellcolor{green!60!black!15}12.1  & \cellcolor{green!60!black!15}6.8   & \cellcolor{green!60!black!15}3.7   \\
(b) = (a) + TTT~\cite{lact}
 & \cellcolor{green!60!black!15}23.12 & \cellcolor{green!60!black!15}0.704 & \cellcolor{green!60!black!15}0.286
 & \cellcolor{green!60!black!55}0.439 & \cellcolor{green!60!black!55}0.862 & \cellcolor{green!60!black!55}1.706
 & \cellcolor{green!60!black!53}49.5  & \cellcolor{green!60!black!53}49.1  & \cellcolor{green!60!black!54}49.1  \\
\midrule
\multicolumn{10}{l}{\textit{TTB (Ours):}} \\
(c) = (b) + context pairing
 & \cellcolor{green!60!black!30}23.78 & \cellcolor{green!60!black!28}0.728 & \cellcolor{green!60!black!28}0.264
 & \cellcolor{green!60!black!50}0.626 & \cellcolor{green!60!black!52}1.237 & \cellcolor{green!60!black!53}2.458
 & \cellcolor{green!60!black!55}50.8  & \cellcolor{green!60!black!55}50.8  & \cellcolor{green!60!black!55}50.7  \\
(d) = (b) + cross-view attn.
 & \cellcolor{green!60!black!38}24.13 & \cellcolor{green!60!black!36}0.742 & \cellcolor{green!60!black!36}0.252
 & \cellcolor{green!60!black!34}1.267 & \cellcolor{green!60!black!23}5.178 & \cellcolor{green!60!black!22}17.745
 & \cellcolor{green!60!black!54}50.4  & \cellcolor{green!60!black!55}50.4  & \cellcolor{green!60!black!55}50.5  \\
(e) = (c) + MSE loss, $T{=}3$
 & \cellcolor{green!60!black!41}24.22 & \cellcolor{green!60!black!39}0.747 & \cellcolor{green!60!black!38}0.248
 & \cellcolor{green!60!black!41}0.994 & \cellcolor{green!60!black!47}1.939 & \cellcolor{green!60!black!51}3.828
 & \cellcolor{green!60!black!55}51.0  & \cellcolor{green!60!black!55}50.6  & \cellcolor{green!60!black!55}50.6  \\
(f) = (e) + cross-view attn.
 & \cellcolor{green!60!black!55}24.84 & \cellcolor{green!60!black!51}0.768 & \cellcolor{green!60!black!49}0.230
 & \cellcolor{green!60!black!15}1.998 & \cellcolor{green!60!black!15}6.251 & \cellcolor{green!60!black!15}20.996
 & \cellcolor{green!60!black!55}50.9  & \cellcolor{green!60!black!55}50.4  & \cellcolor{green!60!black!55}50.5  \\
\bottomrule
\end{tabular}%
g}
\end{table}

%% file: tab/ablation_combine.tex
\begin{table}[t]
\vspace{-2mm}
\begin{minipage}[b]{.48\linewidth}
  \caption{{\bf Inner-loop optimization ablation.}}
  \label{tab:ablation_innerloop}
  \centering
  \setlength{\tabcolsep}{4pt}
  \renewcommand{\arraystretch}{1.0}
  \setlength{\aboverulesep}{1pt}
  \setlength{\belowrulesep}{1pt}
  \resizebox{\linewidth}{!}{%
  \begin{tabular}{lccccc}
    \toprule
    Framework & $T$ & Loss & PSNR$\uparrow$ & SSIM$\uparrow$ & LPIPS$\downarrow$ \\
    \midrule
    TTT~\cite{lact} & 1 & dot-prod & 23.12 & 0.704 & 0.286 \\
    TTT~\cite{lact} & 3 & MSE      & 23.39 & 0.715 & 0.277 \\
    \midrule
    TTB{-}{-} & 1 & dot-prod & 23.78 & 0.728 & 0.264 \\
    TTB{-}{-} & 2 & MSE      & \cellcolor{yellow!25}24.16 & \cellcolor{yellow!25}0.744 & \cellcolor{yellow!25}0.251 \\
    TTB{-}{-} & 3 & MSE      & \cellcolor{orange!25}24.22 & \cellcolor{orange!25}0.747 & \cellcolor{orange!25}0.248 \\
    TTB{-}{-} & 5 & MSE      & \cellcolor{red!25}24.29    & \cellcolor{red!25}0.751    & \cellcolor{red!25}0.246    \\
    \bottomrule
  \end{tabular}%
  }
\end{minipage}%
\hfill
\begin{minipage}[b]{.51\linewidth}
  \caption{{\bf TTB layer design ablation.}}
  \label{tab:ablation_feedback_residual}
  \centering
  \resizebox{\linewidth}{!}{%
  \begin{tabular}{ccccc}
    \toprule
    Feedback & Residual & PSNR$\uparrow$ & SSIM$\uparrow$ & LPIPS$\downarrow$ \\
    \midrule
    \xmark & \xmark & 24.32 & 0.748 & 0.243 \\
    \xmark & \cmark & \cellcolor{yellow!25}24.75 & \cellcolor{orange!25}0.764 & \cellcolor{orange!25}0.233 \\
    \cmark & \xmark & \cellcolor{orange!25}24.76 & \cellcolor{orange!25}0.764 & \cellcolor{yellow!25}0.231 \\
    \cmark & \cmark & \cellcolor{red!25}24.84    & \cellcolor{red!25}0.768    & \cellcolor{red!25}0.230    \\
    \bottomrule
  \end{tabular}%
  }
\end{minipage}
\vspace{-2mm}
\end{table}

%% file: sec/4_0_experiments.tex
\vspace{-2mm}
\section{Experiments}
\label{sec:experiments}
\vspace{-2mm}

%% file: sec/4_1_details.tex
\subsection{Implementation details}
\label{ssec:impl_details}
\vspace{-1mm}
We use 768 latent channels with 12 decoder blocks at patch size~8, replacing each cross-view attention with a TTB layer. We initialize our models from the corresponding DVSM~\citep{dvsm} checkpoint. We use AdamW~\cite{adamw} with cosine annealing and perceptual loss weight $\lambda=0.2$. Models are trained on DL3DV-10K~\cite{dl3dv} dataset with 64 A100 GPUs. We first train at 340p for $100$K iterations with learning rate $4\!\times\!10^{-4}$, then finetune at 480p, 720p, and 960p for $10$K iterations each with learning rate $1\!\times\!10^{-5}$. We randomly sample either $16$ or $32$ context views per scene with the same number of target views. Ablations use a smaller setup on $8$ A100 GPUs at patch size $16$ truncated to 720p, with efficiency metrics still reported at patch size $8$ for consistency. For evaluation, we use the $140$ held-out DL3DV testing scenes and follow previous view sampling~\cite{longlrm}.



\textbf{TTB layer.}
We use a single-headed fast-weight SwiGLU-MLP~\citep{glu} with a hidden dimension of $3072$, a $4\times$ expansion of the model dimension. The fast weight is updated with the Muon optimizer~\citep{muon} and we run $T\!=\!3$ inner-loop update steps. The per-token learning rate is predicted by an additional linear head with base learning rate of $0.01$. The inner-loop objective maps context-rendering tokens to the cross-view attention output of context-ground-truth tokens with the MSE loss. Fast-weight local-duplicate inference (Sec.~\ref{sssec:infer_strategy}) is turned off except stated otherwise. When enabled, we set $B{=}16$. For some experiments we also report a variant denoted TTB{-}{-} or Ours{-}{-}, which replaces the cross-view attention target with a linear projection $W_v X^{(\text{ctx-gt})}$ and thus avoids the quadratic-complexity cross-view attention at reconstruction, making reconstruction time linear in $V$.

%% file: sec/4_2_comparisons.tex
\vspace{-1mm}
\subsection{Comparisons}
\label{ssec:comparisons}

\textbf{DL3DV.} 
Tab.~\ref{tab:dl3dv} and Fig.~\ref{fig:qual_dl3dv} report results on DL3DV with 16, 32, and 64 input views at 960p.
TTB substantially closes the quality gap between DVSM~\cite{dvsm} and LaCT~\cite{lact} while rendering at a constant 23 FPS, up to $23\times$ faster than DVSM and $2.3\times$ faster than LaCT.
Our model is built on DVSM with 12 blocks while LaCT uses 24 blocks (model size in Tab.~\ref{tab:in_the_wild}).
The Ours{-}{-} without cross-view attention is comparable to LaCT quality with less parameters and faster rendering.  
With local-duplicate inference (Sec.~\ref{sssec:infer_strategy}), Ours~(duplicate) further closes the residual gap at large $V$ and approaches DVSM quality at the same rendering speed.
Fig.~\ref{fig:qual_dl3dv} confirms these trends qualitatively, where TTB preserves thin and high-frequency structures such as handrails, window grilles, and small text that LaCT and LVSM~\cite{lvsm} render with visible blur.

\textbf{In-the-wild.}
We further evaluate the generalization ability on out-of-domain datasets with 16 and 64 context views in Tab.~\ref{tab:in_the_wild}.
We keep the aspect ratio of the context images and resize to the resolutions preferred by the models.
The synthesized novel views are resized to match ground-truth source resolutions for evaluation.
Comparisons to LaCT~\cite{lact} and DVSM~\cite{dvsm} are consistent with the in-domain evaluation---our quality is often comparable to DVSM but at a much faster rendering; we use more context prefilling time than LaCT but gaining quality with $2.3\times$ faster rendering.
LagerNVS~\cite{largernvs} is a very recent work which builds upon VGGT~\cite{vggt} and trains on multiple datasets.
We use our intermediate checkpoints at 480p to better align with LagerNVS's input resolution.
Our quality outperforms LagerNVS by a large margin with $3.1$--$9.5\times$ faster rendering.

\textbf{Re10K.}
We additionally compare on Re10K with two low-resolution context views. This is not our main target scenario as the rendering time gain from cross-view attention becomes marginal. We finetune from the Re10K checkpoint of DVSM~\cite{dvsm} with DINO injection. Results are in Tab.~\ref{tab:re10k}. Our TTB can reproduce DVSM cross-attention quality in this setup. See supplementary for full results.

%% file: sec/4_3_abla.tex
\subsection{Ablation studies}
\label{ssec:abla}

\textbf{From TTT to TTB.} Tab.~\ref{tab:ablation_ttb} decomposes the gap between a vanilla LaCT-style TTT replacement~\cite{lact} and our full TTB into three additive components: context pairing, switching to MSE with $T{=}3$ inner steps (yielding TTB{-}{-}), and replacing the linear-projection target with cross-view attention output. Each contributes meaningfully, and stacking all three closes the quality gap to DVSM~\cite{dvsm} while retaining TTT's $\sim\!50$ FPS rendering. The reconstruction cost grows quadratically with $V$ due to the in-loop cross-view attention, consistent with our reconstruction time for rendering quality trade-off.

\textbf{Inner-loop optimization.} Tab.~\ref{tab:ablation_innerloop} ablates the inner-loop loss and number of update steps $T$. Switching from LaCT's unnormalized dot-product loss~\cite{lact} to MSE consistently improves quality for both TTT and TTB{-}{-}, with TTB{-}{-} benefiting more ($+0.44$ vs $+0.27$ PSNR). Increasing $T$ brings monotonic but diminishing gains, leading us to adopt $T{=}3$ with MSE as a sweet spot between quality and reconstruction cost. This asymmetric response aligns with our light field view---TTB's inner loop fits a target function and benefits from convergence, while TTT-KVB is closer to a one-shot linear write.

\textbf{TTB layer design.} Tab.~\ref{tab:ablation_feedback_residual} ablates two design choices in our TTB layer. Residual learning reformulates the inner-loop target from the absolute cross-view attention output to its residual against $X^{(\text{ctx-rend})}$, reducing the magnitude of the regression target and easing the few-step inner-loop optimization. Feedback additionally applies the baked MLP to update the context ground-truth tokens, so that all three token streams propagate through identical operators in subsequent layers. Both designs contribute roughly equally in isolation ($+0.43$ and $+0.44$ PSNR) and combine to $+0.52$.

\textbf{Limitations.}
TTB is constrained by the fixed capacity of its fast-weight MLP, causing quality to saturate at large $V$; the local-duplicate routing that mitigates this is a heuristic, and learning it during training is left for future work.
Its inner-loop cross-view attention also makes reconstruction quadratic in $V$, a drawback for single-shot scenarios.
Finally, TTB is built on reconstruction rather than generation, interpolating within context views without extrapolating to unobserved viewpoints.

%% file: sec/5_conclusion.tex
\vspace{-2mm}
\section{Conclusion}
\vspace{-1mm}
\label{sec:conclusion}
We present Test-Time Baking (TTB), which retargets the TTT inner-loop optimization from raw value tokens to cross-view attention outputs, turning the fast-weight MLP from a replacement for attention into its collaborator. Combined with a context pair that resolves the reconstruction-rendering modality misalignment, TTB bakes DVSM-level scene information into a fixed-size MLP while retaining TTT-level rendering complexity. Across DL3DV, Re10K, and in-the-wild benchmarks, TTB matches DVSM~\cite{dvsm} quality at a constant 23 FPS and consistently surpasses LaCT~\cite{lact}. We hope this baking perspective offers a new way to integrate attention and TTT beyond view synthesis.

%% file: sec/X_pseudocode.tex


\begin{algorithm}
\caption{TTB model training}
\label{algo:ttb_training}
\begin{algorithmic}[1]
\State \textbf{Input:} DVSM~\cite{dvsm} model parameters $\phi$ with $L$ blocks; TTB MLP fast-weight $\{\theta_\ell\}_{\ell=1}^L$; Outer lr $\alpha$; Inner steps $T$; Training set $\mathcal{D}$
\State \textbf{Output:} Optimized model parameters $\phi^*$ and fast weight $\{\theta_\ell^*\}_{l=1}^L$

\State

\Function{InnerOptim}{$\theta^{(0)}, \mathbf{X}, \mathbf{Y}$}
    \State {\color{green!60!black} $\triangleright$ Bake the fast-weight MLP for a $\mathbf{X}\mapsto\mathbf{Y}$ mapping}
    \State \textbf{for} $t = 1, \ldots, T$ \textbf{do}
    \State \hspace{0.5cm} $\mathbf{Y'} = f_{\theta^{(t-1)}}(\mathbf{X})$ \hfill {\color{green!60!black} $\triangleright$ MLP feed-forwarding}
    \State \hspace{0.5cm} $\mathcal{L}_{\text{TTB}} = \|\mathbf{Y'} - \mathbf{Y}\|_2^2$  \hfill {\color{green!60!black} $\triangleright$ Inner-loop objective}
    \State \hspace{0.5cm} {\color{green!60!black} $\triangleright$ We follow LaCT~\cite{lact} to use Muon~\cite{muon} for inner-loop optimization}
    \State \hspace{0.5cm} $\theta^{(t)} = \theta^{(t-1)} - \mathrm{Muon}\left( \nabla_{\theta^{(t-1)}} \mathcal{L}_{\text{TTB}} \right)$  \hfill {\color{green!60!black} $\triangleright$ Fast-weight updating (Eq.\ref{eq:final_ttb})}
    \State \Return $\theta^{(T)}$
\EndFunction

\State

\State \textbf{for} each training iteration \textbf{do}

\State \hspace{0.5cm} {\color{green!60!black} $\triangleright$ Let's assume one scene in a batch for brevity}
\State \hspace{0.5cm} Sample scene: input views $\{(\Image_i, \Pose_i, \Intr_i)\}_{i=1}^N$, target views $\{(\Image'_i, \Pose'_i, \Intr'_i)\}_{i=1}^M$ from $\mathcal{D}$

\State \hspace{0.5cm} {\color{green!60!black} $\triangleright$ Construct the context pairs (Eq.~\ref{eq:ctx_pair}). Omit layer index for brevity.}
\State \hspace{0.5cm} $\tokenctxgt = \text{LayerNorm}(\PEray(\Pose, \Intr) + \PErgb(\Image))$ \hfill {\color{green!60!black} $\triangleright$ Context ground-truth tokens}
\State \hspace{0.5cm} $\tokenctxrend = \text{LayerNorm}(\PEray(\Pose, \Intr))$ \hfill {\color{green!60!black} $\triangleright$ Context rendering tokens}

\State \hspace{0.5cm} {\color{green!60!black} $\triangleright$ Reconstruction stage (context prefilling)}
\State \hspace{0.5cm} \textbf{for} $\ell = 1, \ldots, L$ \textbf{do}
\State \hspace{1cm} {\color{green!60!black} $\triangleright$ Transformer block feed-forwarding (Eq.~\ref{eq:dvsm})}
\State \hspace{1cm} $\tokenctxgt = \tokenctxgt + \text{IntraAttn}_\ell(\tokenctxgt)$ 
\State \hspace{1cm} $\tokenctxrend = \tokenctxrend + \text{IntraAttn}_\ell(\tokenctxrend)$ 
\State \hspace{1cm} $\tokenctxgt = \tokenctxgt + \text{FFN}_{\ell,1}(\tokenctxgt)$
\State \hspace{1cm} $\tokenctxrend = \tokenctxrend + \text{FFN}_{\ell,1}(\tokenctxrend)$

\State \hspace{1cm} {\color{green!60!black} $\triangleright$ {\bf ==== Start of test-time baking ====}}
\State \hspace{1cm} $\mathbf{O} = \text{CrossAttn}_\ell(\tokenctxgt)$
\State \hspace{1cm} $\mathbf{Y} = \text{Linear}_{\ell,v}(\mathbf{O} - \tokenctxrend)$  \hfill {\color{green!60!black} $\triangleright$ The mapping target}
\State \hspace{1cm} $\mathbf{X} = \text{Linear}_{\ell,k}(\tokenctxrend)$ \hfill {\color{green!60!black} $\triangleright$ The mapping source}
\State \hspace{1cm} $\theta_\ell^{(T)} = \text{InnerOptim}(\theta_\ell, \mathbf{X}, \mathbf{Y})$ \hfill {\color{green!60!black} $\triangleright$ Inner-loop optimization for $T$ steps}
\State \hspace{1cm} $\mathbf{Y'} = f_{\theta_\ell^{(T)}}(\mathbf{X})$ \hfill {\color{green!60!black} $\triangleright$ Feed-forward with the baked MLP}
\State \hspace{1cm} $\tokenctxgt = \tokenctxgt + \mathbf{O} + \mathbf{Y'}$ \hfill {\color{green!60!black} $\triangleright$ Apply residual to context ground-truth tokens}
\State \hspace{1cm} $\tokenctxrend = \tokenctxrend + \mathbf{Y'}$  \hfill {\color{green!60!black} $\triangleright$ Apply residual to context rendering tokens}
\State \hspace{1cm} {\color{green!60!black} $\triangleright$ {\bf ==== End of test-time baking ====}}

\State \hspace{1cm} $\tokenctxgt = \tokenctxgt + \text{FFN}_{\ell,2}(\tokenctxgt)$
\State \hspace{1cm} $\tokenctxrend = \tokenctxrend + \text{FFN}_{\ell,2}(\tokenctxrend)$

\State \hspace{0.5cm} {\color{green!60!black} $\triangleright$ Rendering stage}
\State \hspace{0.5cm} $\tokennv = \text{LayerNorm}(\PEray(\Pose', \Intr'))$ \hfill {\color{green!60!black} $\triangleright$ Novel-view rendering tokens}
\State \hspace{0.5cm} \textbf{for} $\ell = 1, \ldots, L$ \textbf{do}
\State \hspace{1cm} $\tokennv = \tokennv + \text{IntraAttn}_\ell(\tokennv)$
\State \hspace{1cm} $\tokennv = \tokennv + \text{FFN}_{\ell,1}(\tokennv)$

\State \hspace{1cm} $\tokennv = \tokennv + f_{\theta_\ell^{(T)}}\left(\text{Linear}_{\ell,k}(\tokennv)\right)$  \hfill {\color{green!60!black} $\triangleright$ {\bf Baked MLPs replace cross-view attentions}}

\State \hspace{1cm} $\tokennv = \tokennv + \text{FFN}_{\ell,2}(\tokennv)$

\State \hspace{0.5cm} $\bar{\Image} = (\text{PixShuf} \circ \text{Sigmoid} \circ \text{Linear} \circ \text{LayerNorm})(\tokennv)$ \hfill {\color{green!60!black} $\triangleright$ Color head (Eq.\ref{eq:rgb_head})}
\State \hspace{0.5cm} $\mathcal{J} = \text{MSE}(\bar{\Image}, \Image') + \lambda \cdot \text{Percep}(\bar{\Image}, \Image')$ \hfill {\color{green!60!black} $\triangleright$ Rendering losses (Eq.\ref{eq:objective})}
\State \hspace{0.5cm} $\phi, \theta = \text{ADAM}(\{\phi, \theta\}, \{\nabla_\phi\mathcal{J}, \nabla_\theta\mathcal{J}\}, \alpha)$  \hfill {\color{green!60!black} $\triangleright$ Optimization step}

\end{algorithmic}
\end{algorithm}

%% file: sec/X_re10k_full.tex
\section{Additional results}
\label{supp:add_results}
\subsection{Few-view lower-resolution results}

We additionally train and evaluate our models on 2-view setup at a lower resolution.
This is not our intended scenario as the rendering FPS speedup by replacing cross-view attentions with our baked MLPs become marginal: 64$\rightarrow$72 at 256$\times$256 and 45$\rightarrow$52 at 512$\times$512.

We follow previous works to train and test on Re10k~\cite{stereomag} dataset. We start from the most capable pretrained checkpoint from DVSM~\cite{dvsm} which employs DINO~\cite{dinov3} injection and patch size of 8. We first train at 256$\times$256 resolution for $100$K iterations with learning rate $4\!\times\!10^{-4}$, and then finetune at 512$\times$512 resolution for $10$K iterations with learning rate $1\!\times\!10^{-4}$. In each iteration on a GPU, we sample $32$ scenes with $2$ context views and $3$ target views. The distance between the $2$ views is randomly sampled between 25--192 frame skips following previous works.

The results comparison is presented in Tab.~\ref{tab:re10k_detail}. Under such few views, our TTBs can faithfully reproduce the cross-view attentions results, as reflected by the comparable results with our pretrained DVSM.

\input{tab/in_the_wild_all}
\subsection{In-the-wild generalization at 16 input views}

We complement the 64-view in-the-wild evaluation in main paper Tab.~\ref{tab:in_the_wild} with a 16-view counterpart in Tab.~\ref{tab:in_the_wild_all}, covering the same three out-of-domain datasets (MipNeRF360~\cite{mip-nerf360}, Free~\cite{f2nerf}, Hike~\cite{hike}) at both 960p and 512-pixel longer-side settings. The trends mirror our in-domain DL3DV results: TTB closely matches DVSM~\cite{dvsm} quality across all three datasets while consistently surpassing LaCT~\cite{lact} and LVSM~\cite{lvsm}, and outperforms LagerNVS~\cite{largernvs} by a clear margin under the 512-pixel setting. Efficiency-wise, TTB retains its constant-FPS rendering at 16 views just as at 64 views, confirming that the rendering speedup is independent of context view count. These results indicate that TTB's quality and speed advantages generalize beyond the dense-view regime to fewer-view, out-of-domain scenarios.

\input{tab/ablation_chunk}
\subsection{Inference at scale on DL3DV.}
A single fast-weight MLP per TTB layer has fixed capacity, which becomes a bottleneck as the number of context views $V$ grows.
Tab.~\ref{tab:ablation_chunk} compares this single-MLP baseline against our local-duplicate variant, which maintains $K$ local copies of the per-layer fast-weight state, each updated only by views routed to its spatially-local group via k-means over per-view camera features (results reported with our ablation model at 720p).
Up to $V{=}32$, the two variants are indistinguishable and both match DVSM~\cite{dvsm}.
Beyond $V{=}64$, the single-MLP variant saturates and plateaus around $25.8$ PSNR, while the local-duplicate variant continues to scale and reaches $27.74$ at $V{=}256$, on par with DVSM's $27.56$.
Throughout, rendering FPS stays in the $67$ to $72$ range, in stark contrast to DVSM's drop from $63.6$ at $V{=}16$ to $12.6$ at $V{=}256$.
Local duplication thus resolves the capacity ceiling of a single fast-weight MLP without compromising the constant-complexity rendering that motivates TTB.

\input{tab/capacity}
\subsection{Capacity scaling}
\label{supp:capacity}
Tab.~\ref{tab:capacity} studies the two knobs that set the capacity of the baked state. Sweeping $K$ jointly with $V$ shows that the best $K$ is always $V/16$, so local duplication is governed by a single scene-independent budget of $B{=}16$ views per copy rather than a per-scene heuristic; we use this budget for all \emph{Ours (duplicate)} entries in Tab.~\ref{tab:dl3dv}, and the $K{=}1$ column and the $K{=}V/16$ diagonal reproduce the two TTB rows of Tab.~\ref{tab:ablation_chunk}. Widening the fast-weight MLP closes the gap as well and reaches DVSM at $16{\times}$ width, indicating that the remaining gap is a capacity limit rather than a limitation of baking. Since width enlarges the model and requires retraining, we keep the $4{\times}$ width and scale capacity through $K$.

\begin{table}[tb]
  \caption{{\bf Pretraining ablation.}}
  \label{tab:ablation_pretrain}
  \centering
  \setlength{\tabcolsep}{4pt}
  \begin{adjustbox}{max width=\linewidth}
  \begin{tabular}{lcccccc}
    \toprule
    \multirow{2}{*}{Method} & \multicolumn{3}{c}{w/o pretrain} & \multicolumn{3}{c}{w/ pretrain} \\
    \cmidrule(lr){2-4} \cmidrule(lr){5-7}
    & PSNR$\uparrow$ & SSIM$\uparrow$ & LPIPS$\downarrow$ & PSNR$\uparrow$ & SSIM$\uparrow$ & LPIPS$\downarrow$ \\
    \midrule
    TTT & 22.84 & 0.688 & 0.307 & 23.12 & 0.704 & 0.286 \\
    TTB{-}{-}       & 23.51 & 0.720 & 0.274 & 24.22 & 0.747 & 0.248 \\
    TTB             & \textbf{23.82} & \textbf{0.731} & \textbf{0.264} & \textbf{24.84} & \textbf{0.768} & \textbf{0.230} \\
    \bottomrule
  \end{tabular}
  \end{adjustbox}
\end{table}

\subsection{Effect of pretraining.} Tab.~\ref{tab:ablation_pretrain} compares three frameworks trained from scratch versus initialized from a DVSM~\cite{dvsm} checkpoint (results reported with our ablation model at 720p). The PSNR gain from pretraining grows monotonically with cross-view attention preservation: smallest for TTT (no cross-view attention), larger for TTB{-}{-} (linear target), and largest for our full TTB (cross-view attention target), confirming that preserving cross-view attention is key to re-using decoder-only pretrained weights. Notably, TTB from scratch already surpasses pretrained TTT, indicating that its advantage is structural.

\definecolor{effgreen}{RGB}{216, 238, 216}
\definecolor{bestred}{RGB}{255, 195, 195}
\definecolor{secondorange}{RGB}{255, 218, 180}

\begin{table}[t]
\centering
\caption{\textbf{Efficiency comparison on DL3DV at $544{\times}960$ resolution with 32 input views.}
\underline{Recon-time} is the processing time of a single scene.
\underline{Rend-fps} indicates the number of frames rendered per second without batching.
\underline{Infer-mem} is the peak GPU memory usage when measuring the reconstruction and rendering.
\underline{Train-time/it} is the sum of model forward and backward time per iteration under training mode with a batch containing 32 context views.
\underline{Train-mem} is the peak memory usage when measuring training time.
The actual training randomly samples different number of input views per scene rather than a fixed number.
We \colorbox{effgreen}{highlight} cells with numbers similar to the most efficient one under each column.}
\label{tab:efficiency}
\setlength{\tabcolsep}{3pt}
\small
\resizebox{\linewidth}{!}{%
\begin{tabular}{llccc|cccccc}
\toprule
Method & Venue & PSNR$\uparrow$ & SSIM$\uparrow$ & LPIPS$\downarrow$ & \#params$\downarrow$ & Recon-time$\downarrow$ & Rend-fps$\uparrow$ & Infer-mem$\downarrow$ & Train-time/it$\downarrow$ & Train-mem$\downarrow$ \\
\midrule
LVSM~\cite{lvsm}                  & ICLR'25 & 22.24 & 0.691 & 0.352 & \cellcolor{effgreen}165M & 15s  & 20 & \cellcolor{effgreen}9G  & 73s   & 42G \\
LaCT~\cite{lact}                  & ICLR'26 & 26.90 & 0.837 & 0.185 & 379M                     & 3.7s & 10 & 23G                     & 36s   & 72G \\
DVSM $_{\text{ps8}}$        & arXiv'26      & \cellcolor{bestred}29.13 & \cellcolor{bestred}0.893 & \cellcolor{bestred}0.129 & \cellcolor{effgreen}163M & 15s  & 2  & 20G & 133s & 65G \\
\midrule
TTB-- (ours)                      & --      & 28.12 & 0.867 & 0.158 & 244M & 3.7s & \cellcolor{effgreen}23 & 18G & 26s  & 68G \\
TTB (ours)                        & --      & \cellcolor{secondorange}28.92 & \cellcolor{secondorange}0.882 & \cellcolor{secondorange}0.145 & 271M & 18s  & \cellcolor{effgreen}23 & 18G & 90s  & 76G \\
\bottomrule
\end{tabular}%
}
\end{table}

\subsection{Efficiency analysis}
\label{sec:efficiency}

We provide a detailed efficiency comparison in Tab.~\ref{tab:efficiency}, reporting parameter counts, per-scene reconstruction time, per-view rendering FPS, and peak GPU memory at both inference and training time. All numbers are measured at $544{\times}960$ resolution with $V{=}32$ input views.

\input{tab/end2end}
\paragraph{End-to-end time.} Since TTB shifts computation from rendering to reconstruction, Tab.~\ref{tab:end2end} reports the total time to reconstruct a scene and render $N$ novel views. TTB overtakes DVSM after only 5--7 views at every $V$, a threshold that any interactive viewer or trajectory rendering exceeds, while TTB{-}{-} leads from the first view at the quality cost reported in Tab.~\ref{tab:dl3dv}.

\subsection{Scene representation size}
\label{sec:scene-rep-size}

\begin{figure}[!htb]
\centering
\includegraphics[width=0.7\linewidth]{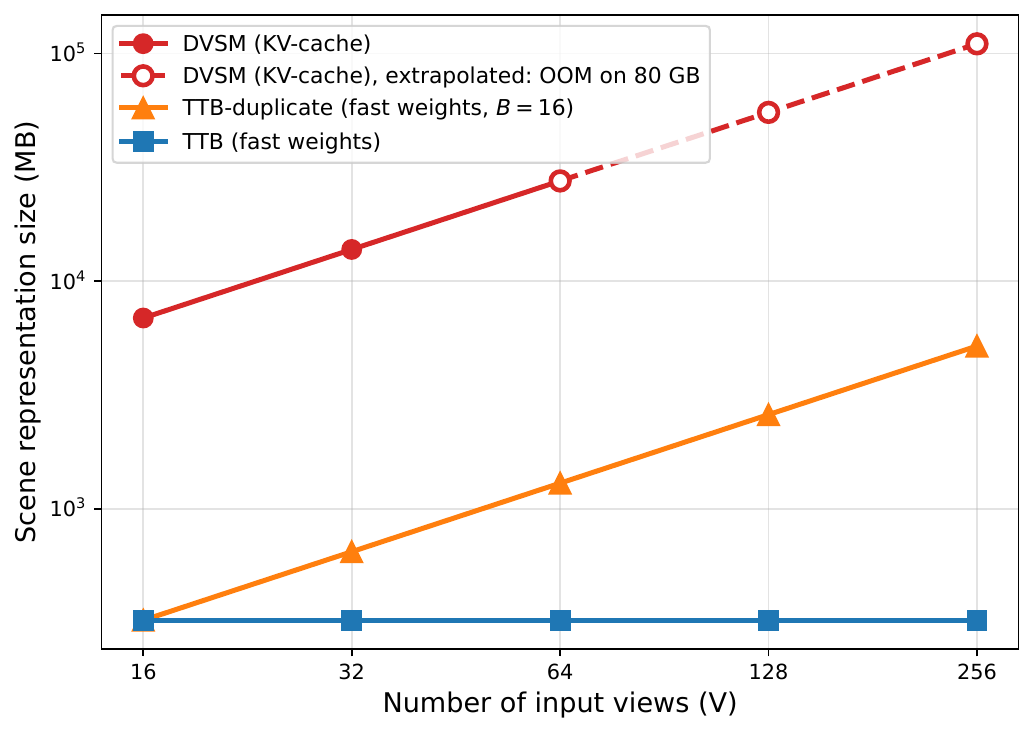}
\caption{\textbf{Scene representation size as $V$ grows} at $544{\times}960$ resolution, measured on a single A100. The DVSM~\cite{dvsm} KV-cache grows linearly in $V$ (hollow points: linear extrapolation where DVSM ran out of 80\,GB memory), TTB's baked fast weights stay constant, and TTB-duplicate grows with the number of copies $K{=}V/16$ while remaining $21{\times}$ smaller than the KV-cache.}
\label{fig:scene-rep-size}
\end{figure}

We further quantify the memory footprint of the per-scene state that must be retained between reconstruction and rendering at $544{\times}960$ resolution. Fig.~\ref{fig:scene-rep-size} compares the KV-cache of DVSM~\cite{dvsm} against the baked fast-weight state of TTB as the number of context views $V$ grows. The DVSM KV-cache scales linearly with $V$, reaching tens of GBs at $V{=}256$, whereas TTB's scene representation is a fixed-size set of fast-weight matrices whose footprint is independent of $V$ ($324$\,MB across all $L$ blocks in FP32). At $V{=}64$ this is an $85{\times}$ reduction over DVSM, growing to $340{\times}$ at $V{=}256$; local-duplicate inference stores $K{=}V/16$ copies and stays a constant $21{\times}$ smaller than the KV-cache at every $V$. This compact, $V$-independent state is what enables TTB's $O(1)$ per-token rendering and is also a practical advantage for deployment scenarios with limited GPU memory.

%% file: tab/in_the_wild_all.tex
\begin{table}[t]
\centering
\caption{{\bf In-the-wild generalization at 16 input views.}}
\label{tab:in_the_wild_all}
\begin{adjustbox}{max width=\linewidth}
\begin{tabular}{
    l@{\hskip 3pt}c
    ccc c 
    ccc c 
    ccc c  cc
}
    \toprule
    &&
    \multicolumn{3}{c}{\bf MipNerf360~\cite{mip-nerf360}} &&
    \multicolumn{3}{c}{\bf Free~\cite{f2nerf}} &&
    \multicolumn{3}{c}{\bf Hike~\cite{hike}} &&
    \multicolumn{2}{c}{\bf Efficiency} \\
    \cmidrule(lr){3-5}\cmidrule(lr){7-9}\cmidrule(lr){11-13}\cmidrule(lr){15-16}
    Method & Size &
    PSNR$\uparrow$&SSIM$\uparrow$&LPIPS$\downarrow$ &&
    PSNR$\uparrow$&SSIM$\uparrow$&LPIPS$\downarrow$ &&
    PSNR$\uparrow$&SSIM$\uparrow$&LPIPS$\downarrow$ &&
    Time$\downarrow$ & FPS$\uparrow$\\
    \midrule
    \multicolumn{14}{@{}l@{}}{\bf \underline{Input: 16 views $\times$ 518{,}400 pixels per view (960p)}} \\
    LVSM & {\scriptsize 165M} &
    19.58 & 0.459 & 0.564 &&
    19.95 & 0.528 & 0.441 &&
    18.48 & 0.454 & 0.492 &&
    \cellcolor{yellow!25}3.8s & \cellcolor{yellow!25}20 \\
    LaCT & {\scriptsize 379M} &
    20.10 & 0.494 & 0.498 &&
    20.77 & 0.588 & 0.350 &&
    18.35 & 0.512 & 0.403 &&
    \cellcolor{red!25}1.9s & 10 \\
    DVSM & {\scriptsize 163M} &
    \cellcolor{orange!25}21.51 & \cellcolor{orange!25}0.553 & \cellcolor{orange!25}0.427 &&
    \cellcolor{yellow!25}22.04 & \cellcolor{orange!25}0.664 & \cellcolor{red!25}0.285 &&
    \cellcolor{red!25}20.24 & \cellcolor{red!25}0.591 & \cellcolor{red!25}0.363 &&
    4.4s & 4 \\
    Ours{-}{-} & {\scriptsize 244M} &
    \cellcolor{yellow!25}21.46 & \cellcolor{yellow!25}0.540 & \cellcolor{yellow!25}0.442 &&
    \cellcolor{orange!25}22.21 & \cellcolor{yellow!25}0.638 & \cellcolor{yellow!25}0.321 &&
    \cellcolor{yellow!25}19.25 & \cellcolor{yellow!25}0.549 & \cellcolor{yellow!25}0.391 &&
    \cellcolor{red!25}1.9s & \cellcolor{red!25}23 \\
    Ours & {\scriptsize 271M} &
    \cellcolor{red!25}22.03 & \cellcolor{red!25}0.563 & \cellcolor{red!25}0.424 &&
    \cellcolor{red!25}22.93 & \cellcolor{red!25}0.673 & \cellcolor{red!25}0.285 &&
    \cellcolor{orange!25}19.94 & \cellcolor{orange!25}0.585 & \cellcolor{orange!25}0.368 &&
    5.4s & \cellcolor{red!25}23 \\
    \midrule
    \multicolumn{14}{@{}l@{}}{\bf \underline{Input: 16 views $\times$ 512 pixels at image longer side}} \\
    LagerNVS & {\scriptsize 1052M} &
    20.41 & 0.469 & 0.523 &&
    19.33 & 0.490 & 0.412 &&
    17.98 & 0.426 & 0.457 &&
    \cellcolor{red!25}0.4s & 21 \\
    DVSM & {\scriptsize 163M} &
    \cellcolor{orange!25}21.47 & \cellcolor{red!25}0.531 & \cellcolor{red!25}0.482 &&
    \cellcolor{orange!25}21.71 & \cellcolor{red!25}0.637 & \cellcolor{red!25}0.336 &&
    \cellcolor{red!25}20.23 & \cellcolor{red!25}0.586 & \cellcolor{red!25}0.347 &&
    \cellcolor{red!25}0.4s & \cellcolor{orange!25}33 \\
    Ours{-}{-} & {\scriptsize 244M} & 
    \cellcolor{yellow!25}21.45 & \cellcolor{yellow!25}0.519 & \cellcolor{yellow!25}0.501 &&
    \cellcolor{yellow!25}21.65 & \cellcolor{yellow!25}0.605 & \cellcolor{yellow!25}0.374 &&
    \cellcolor{yellow!25}19.26 & \cellcolor{yellow!25}0.542 & \cellcolor{yellow!25}0.380 &&
    \cellcolor{yellow!25}0.5s & \cellcolor{red!25}67 \\
    Ours & {\scriptsize 271M} &
    \cellcolor{red!25}21.75 & \cellcolor{red!25}0.531 & \cellcolor{orange!25}0.486 &&
    \cellcolor{red!25}22.36 & \cellcolor{red!25}0.637 & \cellcolor{orange!25}0.346 &&
    \cellcolor{orange!25}19.92 & \cellcolor{orange!25}0.572 & \cellcolor{orange!25}0.361 &&
    0.8s & \cellcolor{red!25}67 \\
    \bottomrule
\end{tabular}
\end{adjustbox}
\end{table}

%% file: tab/ablation_chunk.tex

\begin{table}[t]
\centering
\small
\setlength{\tabcolsep}{6pt}
\caption{\textbf{Inference at scale on DL3DV.}
As the number of context views $V$ grows, \emph{TTB-single}---a single fast-weight MLP per TTB layer that absorbs all $V$ views---suffers \textbf{capacity saturation}: its PSNR plateaus around $V{=}64$ and even regresses thereafter.
\emph{TTB-duplicate}, which maintains $K$ local copies of the per-layer fast-weight state, each baked from a spatially-local group of views, prevents this saturation and tracks DVSM's quality scaling while preserving $\mathcal{O}(1)$ per-frame rendering.
Best per column is shaded red, second-best orange.}
\label{tab:ablation_chunk}
\begin{tabular}{l l ccccc}
\toprule
& Method & $V{=}16$ & $V{=}32$ & $V{=}64$ & $V{=}128$ & $V{=}256$ \\
\midrule
\multirow{3}{*}{PSNR$\uparrow$}
& DVSM          & \cellcolor{orange!25}22.83 & \cellcolor{orange!25}24.68 & \cellcolor{red!25}26.23    & \cellcolor{red!25}27.20    & \cellcolor{orange!25}27.56 \\
& TTB-single    & \cellcolor{red!25}23.26    & \cellcolor{red!25}24.84    & 25.60                      & 25.82                      & 25.78 \\
& TTB-duplicate & \cellcolor{red!25}23.26    & \cellcolor{red!25}24.84    & \cellcolor{orange!25}26.13 & \cellcolor{orange!25}27.14 & \cellcolor{red!25}27.74 \\
\midrule
\multirow{3}{*}{Recon-time$\downarrow$}
& DVSM          & \cellcolor{red!25}0.136    & \cellcolor{red!25}0.389    & \cellcolor{red!25}1.258    & \cellcolor{red!25}4.815    & \cellcolor{red!25}18.106 \\
& TTB-single    & 0.337                      & \cellcolor{orange!25}0.740 & \cellcolor{orange!25}1.911 & \cellcolor{orange!25}6.065 & \cellcolor{orange!25}20.549 \\
& TTB-duplicate & \cellcolor{orange!25}0.333 & 1.048                      & 3.578                      & 13.557                     & 52.064 \\
\midrule
\multirow{3}{*}{Rend-fps$\uparrow$}
& DVSM          & 63.6                    & 61.9                    & 40.1                    & 23.2                    & 12.6 \\
& TTB-single    & \cellcolor{red!25}73.2  & \cellcolor{red!25}72.6  & \cellcolor{red!25}72.3  & \cellcolor{red!25}72.2  & \cellcolor{red!25}72.2 \\
& TTB-duplicate & \cellcolor{orange!25}72.0 & \cellcolor{orange!25}67.9 & \cellcolor{orange!25}68.0 & \cellcolor{orange!25}67.0 & \cellcolor{orange!25}68.5 \\
\bottomrule
\end{tabular}
\end{table}

%% file: tab/capacity.tex
\begin{table}[t]
\centering
\small
\setlength{\tabcolsep}{5pt}
\caption{\textbf{Capacity scaling.} Left: PSNR of local-duplicate inference over the number of copies $K$ and context views $V$ (ablation setting, 720p); every row peaks at $K{=}V/16$, i.e.\ a fixed budget of 16 views per copy (bold). Right: scaling the fast-weight MLP width instead (DL3DV, $V{=}64$, 340p); quality rises monotonically and a $16{\times}$ width matches DVSM.}
\label{tab:capacity}
\begin{adjustbox}{max width=\linewidth}
\begin{tabular}{c ccccc}
\toprule
$V \backslash K$ & 1 & 2 & 4 & 8 & 16 \\
\midrule
16  & \textbf{23.26} & 22.86 & 22.07 & 20.73 & 19.02 \\
32  & \textbf{24.84} & \textbf{24.84} & 24.44 & 23.53 & 22.00 \\
64  & 25.60 & 26.11 & \textbf{26.13} & 25.76 & 24.93 \\
128 & 25.82 & 26.60 & 27.07 & \textbf{27.14} & 26.85 \\
256 & 25.78 & 26.64 & 27.26 & 27.66 & \textbf{27.74} \\
\bottomrule
\end{tabular}
\hspace{2em}
\begin{tabular}{l c ccc}
\toprule
Method & Width & PSNR$\uparrow$ & SSIM$\uparrow$ & LPIPS$\downarrow$ \\
\midrule
DVSM                   & --    & 25.60 & 0.801 & 0.181 \\
TTB ($2{\times}$)      & 1536  & 25.18 & 0.773 & 0.203 \\
TTB ($4{\times}$, ours)& 3072  & 25.36 & 0.780 & 0.197 \\
TTB ($8{\times}$)      & 6144  & 25.52 & 0.786 & 0.192 \\
TTB ($16{\times}$)     & 12288 & 25.61 & 0.792 & 0.188 \\
\bottomrule
\end{tabular}
\end{adjustbox}
\end{table}

%% file: tab/end2end.tex
\begin{table}[t]
\centering
\small
\setlength{\tabcolsep}{5pt}
\caption{\textbf{End-to-end time for rendering $N$ novel views} (seconds, DL3DV at 960p), computed as reconstruction time plus $N/\mathrm{FPS}$ from Tab.~\ref{tab:dl3dv}. TTB overtakes DVSM after 5--7 views at every $V$, while TTB{-}{-} leads from $N{=}1$.}
\label{tab:end2end}
\begin{tabular}{c l cc cccccc}
\toprule
$V$ & Method & Recon & FPS & $N{=}1$ & $N{=}2$ & $N{=}5$ & $N{=}10$ & $N{=}30$ & $N{=}100$ \\
\midrule
\multirow{3}{*}{16}
& DVSM      & 4.4  & 4  & 4.7  & 4.9  & 5.7  & 6.9  & 11.9 & 29.4 \\
& TTB{-}{-} & 1.9  & 23 & 1.9  & 2.0  & 2.1  & 2.3  & 3.2  & 6.2 \\
& TTB       & 5.4  & 23 & 5.4  & 5.5  & 5.6  & 5.8  & 6.7  & 9.7 \\
\midrule
\multirow{3}{*}{32}
& DVSM      & 15.0 & 2  & 15.5 & 16.0 & 17.5 & 20.0 & 30.0 & 65.0 \\
& TTB{-}{-} & 3.7  & 23 & 3.7  & 3.8  & 3.9  & 4.1  & 5.0  & 8.0 \\
& TTB       & 18.0 & 23 & 18.0 & 18.1 & 18.2 & 18.4 & 19.3 & 22.3 \\
\midrule
\multirow{3}{*}{64}
& DVSM      & 60.0 & 1  & 61.0 & 62.0 & 65.0 & 70.0 & 90.0 & 160.0 \\
& TTB{-}{-} & 7.5  & 23 & 7.5  & 7.6  & 7.7  & 7.9  & 8.8  & 11.8 \\
& TTB       & 65.0 & 23 & 65.0 & 65.1 & 65.2 & 65.4 & 66.3 & 69.3 \\
\bottomrule
\end{tabular}
\end{table}

%% file: tab/re10k_detail.tex
\begin{table}[tb]
  \caption{{\bf Two input views on Re10K.}
  }
  \label{tab:re10k_detail}
  \centering
  \begin{adjustbox}{max width=\linewidth}
  \begin{tabular}{l@{}r@{}c ccc}
  \toprule
    Method & Venue && PSNR$\uparrow$ & SSIM$\uparrow$ & LPIPS$\downarrow$ \\
  \midrule
    \multicolumn{5}{@{}l}{\bf \underline{Resolution: 256$\times$256}} \\
    pixelNeRF~\cite{pixelnerf} & CVPR'21
    && 20.43 & 0.589 & 0.550 \\
    AttnRend~\cite{attnrend} & CVPR'23
    && 24.78 & 0.820 & 0.213 \\
    ViewCrafter~\cite{viewcrafter} & TPAMI'24
    && 21.63 & 0.642 & 0.175 \\
    GPNR~\cite{gpnr} & ECCV'24
    && 24.11 & 0.793 & 0.255 \\
    PixelSplat~\cite{pixelsplat} & CVPR'24
    && 26.09 & 0.863 & 0.136 \\
    MVSplat~\cite{mvsplat} & ECCV'24
    && 26.39 & 0.869 & 0.128 \\
    GS-LRM~\cite{gslrm} & ECCV'24
    && 28.10 & 0.892 & 0.114 \\
    DepthSplat~\cite{depthsplat} & CVPR'25
    && 27.46 & 0.889 & 0.115 \\
    SEVA~\cite{seva} & ICCV'25
    && 25.66 & 0.841 & 0.139 \\
    LVSM~\cite{lvsm} & ICLR'25
    && 28.58 & 0.893 & 0.114 \\
    Dec.-only~\cite{lvsm} & ICLR'25
    && 29.67 & 0.906 & 0.098 \\
    Btimer~\cite{btimer} & NeurIPS'25
    && 26.49 & 0.886 & 0.096 \\
    Eff. LVSM~\cite{efflvsm} & ICLR'26
    && 28.93 & 0.895 & 0.102 \\
    TokenGS~\cite{tokengs} & CVPR'26
    && 28.41 & 0.903 & 0.135 \\
    SVSM~\cite{svsm} & CVPR'26
    && 30.01 & 0.910 & 0.096 \\
    DVSM~\cite{dvsm} & arXiv'26
    && \cellcolor{orange!25}31.23 & \cellcolor{orange!25}0.925 & \cellcolor{orange!25}0.085 \\
    Ours{-}{-} &
    && \cellcolor{yellow!25}31.04 & \cellcolor{yellow!25}0.921 & \cellcolor{yellow!25}0.086 \\
    Ours &
    && \cellcolor{red!25}31.35 & \cellcolor{red!25}0.926 & \cellcolor{red!25}0.083 \\
  \midrule
    \multicolumn{5}{@{}l}{\bf \underline{Resolution: 512$\times$512}} \\
    LVSM~\cite{lvsm} & ICLR'25
    && 28.55 & 0.894 & 0.173 \\
    Dec.-only~\cite{lvsm} & ICLR'25
    && 29.53 & 0.904 & 0.141 \\
    Eff. LVSM~\cite{efflvsm} & ICLR'26
    && 29.86 & 0.905 & 0.147 \\
    DVSM~\cite{dvsm} & arXiv'26
    && \cellcolor{orange!25}31.17 & \cellcolor{orange!25}0.924 & \cellcolor{orange!25}0.129 \\
    Ours{-}{-} &
    && \cellcolor{yellow!25}31.00 & \cellcolor{yellow!25}0.921 & \cellcolor{yellow!25}0.130 \\
    Ours &
    && \cellcolor{red!25}31.28 & \cellcolor{red!25}0.926 & \cellcolor{red!25}0.127 \\
  \bottomrule
  \end{tabular}
  \end{adjustbox}
\end{table}

%% file: neurips_2026.bbl
\begin{thebibliography}{79}
\providecommand{\natexlab}[1]{#1}
\providecommand{\url}[1]{\texttt{#1}}
\expandafter\ifx\csname urlstyle\endcsname\relax
  \providecommand{\doi}[1]{doi: #1}\else
  \providecommand{\doi}{doi: \begingroup \urlstyle{rm}\Url}\fi

\bibitem[Asim et~al.(2026)Asim, Wewer, and Lenssen]{scenetok}
Mohammad Asim, Christopher Wewer, and Jan~Eric Lenssen.
\newblock Scenetok: {A} compressed, diffusable token space for 3d scenes.
\newblock In \emph{CVPR}, 2026.

\bibitem[Barron et~al.(2021)Barron, Mildenhall, Tancik, Hedman, Martin{-}Brualla, and Srinivasan]{mipnerf}
Jonathan~T. Barron, Ben Mildenhall, Matthew Tancik, Peter Hedman, Ricardo Martin{-}Brualla, and Pratul~P. Srinivasan.
\newblock Mip-nerf: {A} multiscale representation for anti-aliasing neural radiance fields.
\newblock In \emph{ICCV}, 2021.

\bibitem[Barron et~al.(2022)Barron, Mildenhall, Verbin, Srinivasan, and Hedman]{mip-nerf360}
Jonathan~T. Barron, Ben Mildenhall, Dor Verbin, Pratul~P. Srinivasan, and Peter Hedman.
\newblock Mip-nerf 360: Unbounded anti-aliased neural radiance fields.
\newblock In \emph{CVPR}, 2022.

\bibitem[Barron et~al.(2023)Barron, Mildenhall, Verbin, Srinivasan, and Hedman]{zipnerf}
Jonathan~T. Barron, Ben Mildenhall, Dor Verbin, Pratul~P. Srinivasan, and Peter Hedman.
\newblock Zip-nerf: Anti-aliased grid-based neural radiance fields.
\newblock In \emph{ICCV}, 2023.

\bibitem[Chan et~al.(2021)Chan, Lin, Chan, Nagano, Pan, Mello, Gallo, Guibas, Tremblay, Khamis, Karras, and Wetzstein]{triplane}
Eric Chan, Connor~Z. Lin, Matthew~A. Chan, Koki Nagano, Boxiao Pan, Shalini~De Mello, Orazio Gallo, Leonidas~J. Guibas, Jonathan Tremblay, S.~Khamis, Tero Karras, and Gordon Wetzstein.
\newblock Efficient geometry-aware 3d generative adversarial networks.
\newblock In \emph{CVPR}, 2021.

\bibitem[Charatan et~al.(2024)Charatan, Li, Tagliasacchi, and Sitzmann]{pixelsplat}
David Charatan, Sizhe~Lester Li, Andrea Tagliasacchi, and Vincent Sitzmann.
\newblock Pixelsplat: 3d gaussian splats from image pairs for scalable generalizable 3d reconstruction.
\newblock In \emph{CVPR}, 2024.

\bibitem[Chen et~al.(2021)Chen, Xu, Zhao, Zhang, Xiang, Yu, and Su]{mvsnerf}
Anpei Chen, Zexiang Xu, Fuqiang Zhao, Xiaoshuai Zhang, Fanbo Xiang, Jingyi Yu, and Hao Su.
\newblock Mvsnerf: Fast generalizable radiance field reconstruction from multi-view stereo.
\newblock In \emph{ICCV}, 2021.

\bibitem[Chen et~al.(2022)Chen, Xu, Geiger, Yu, and Su]{tensorf}
Anpei Chen, Zexiang Xu, Andreas Geiger, Jingyi Yu, and Hao Su.
\newblock Tensorf: Tensorial radiance fields.
\newblock In \emph{ECCV}, 2022.

\bibitem[Chen et~al.(2025{\natexlab{a}})Chen, Chen, Xiu, Geiger, and Chen]{ttt3r}
Xingyu Chen, Yue Chen, Yuliang Xiu, Andreas Geiger, and Anpei Chen.
\newblock {TTT3R:} 3d reconstruction as test-time training.
\newblock In \emph{ICLR}, 2025{\natexlab{a}}.

\bibitem[Chen et~al.(2024)Chen, Xu, Zheng, Zhuang, Pollefeys, Geiger, Cham, and Cai]{mvsplat}
Yuedong Chen, Haofei Xu, Chuanxia Zheng, Bohan Zhuang, Marc Pollefeys, Andreas Geiger, Tat{-}Jen Cham, and Jianfei Cai.
\newblock Mvsplat: Efficient 3d gaussian splatting from sparse multi-view images.
\newblock In \emph{ECCV}, 2024.

\bibitem[Chen et~al.(2025{\natexlab{b}})Chen, Tan, Wang, Xu, and Li]{longlrmpp}
Ziwen Chen, Hao Tan, Peng Wang, Zexiang Xu, and Fuxin Li.
\newblock Long-lrm++: Preserving fine details in feed-forward wide-coverage reconstruction.
\newblock \emph{arXiv preprint arXiv:2512.10267}, 2025{\natexlab{b}}.

\bibitem[Chen et~al.(2025{\natexlab{c}})Chen, Tan, Zhang, Bi, Luan, Hong, Li, and Xu]{longlrm}
Ziwen Chen, Hao Tan, Kai Zhang, Sai Bi, Fujun Luan, Yicong Hong, Fuxin Li, and Zexiang Xu.
\newblock Long-lrm: Long-sequence large reconstruction model for wide-coverage gaussian splats.
\newblock In \emph{ICCV}, 2025{\natexlab{c}}.

\bibitem[Du et~al.(2023)Du, Smith, Tewari, and Sitzmann]{attnrend}
Yilun Du, Cameron Smith, Ayush Tewari, and Vincent Sitzmann.
\newblock Learning to render novel views from wide-baseline stereo pairs.
\newblock In \emph{CVPR}, 2023.

\bibitem[Elflein et~al.(2026)Elflein, Li, Agostinho, Gojcic, Leal{-}Taix{\'{e}}, Zhou, and Osep]{vggttt}
Sven Elflein, Ruilong Li, S{\'{e}}rgio Agostinho, Zan Gojcic, Laura Leal{-}Taix{\'{e}}, Qunjie Zhou, and Aljosa Osep.
\newblock Vgg-t\({}^{\mbox{3}}\): Offline feed-forward 3d reconstruction at scale.
\newblock In \emph{CVPR}, 2026.

\bibitem[Fridovich{-}Keil et~al.(2022)Fridovich{-}Keil, Yu, Tancik, Chen, Recht, and Kanazawa]{plenoxels}
Sara Fridovich{-}Keil, Alex Yu, Matthew Tancik, Qinhong Chen, Benjamin Recht, and Angjoo Kanazawa.
\newblock Plenoxels: Radiance fields without neural networks.
\newblock In \emph{CVPR}, 2022.

\bibitem[Gu et~al.(2019)Gu, Fan, Zhu, Dai, Tan, and Tan]{casmvsnet}
Xiaodong Gu, Zhiwen Fan, Siyu Zhu, Zuozhuo Dai, Feitong Tan, and Ping Tan.
\newblock Cascade cost volume for high-resolution multi-view stereo and stereo matching.
\newblock In \emph{CVPR}, 2019.

\bibitem[Guo et~al.(2019)Guo, Yang, Yang, Wang, and Li]{gwcnet}
Xiaoyang Guo, Kai Yang, Wukui Yang, Xiaogang Wang, and Hongsheng Li.
\newblock Group-wise correlation stereo network.
\newblock In \emph{CVPR}, 2019.

\bibitem[Han et~al.(2026)Han, Li, Li, Cao, Wang, Song, Cheng, Zheng, and Huang]{vittt}
Dongchen Han, Yining Li, Tianyu Li, Zixuan Cao, Ziming Wang, Jun Song, Yu~Cheng, Bo~Zheng, and Gao Huang.
\newblock Vit\({}^{\mbox{3}}\): Unlocking test-time training in vision.
\newblock In \emph{CVPR}, 2026.

\bibitem[He et~al.(2016)He, Zhang, Ren, and Sun]{resnet}
Kaiming He, Xiangyu Zhang, Shaoqing Ren, and Jian Sun.
\newblock Deep residual learning for image recognition.
\newblock In \emph{CVPR}, 2016.

\bibitem[Henry et~al.(2020)Henry, Dachapally, Pawar, and Chen]{qknorm}
Alex Henry, Prudhvi~Raj Dachapally, Shubham~Shantaram Pawar, and Yuxuan Chen.
\newblock Query-key normalization for transformers.
\newblock In \emph{EMNLP}, 2020.

\bibitem[Hong et~al.(2024)Hong, Zhang, Gu, Bi, Zhou, Liu, Liu, Sunkavalli, Bui, and Tan]{lrm}
Yicong Hong, Kai Zhang, Jiuxiang Gu, Sai Bi, Yang Zhou, Difan Liu, Feng Liu, Kalyan Sunkavalli, Trung Bui, and Hao Tan.
\newblock {LRM:} large reconstruction model for single image to 3d.
\newblock In \emph{ICLR}, 2024.

\bibitem[Jia et~al.(2026)Jia, Sun, You, Wong, Zou, Yan, Wu, and Jiang]{efflvsm}
Xiaosong Jia, Yihang Sun, Junqi You, Songbur Wong, Zichen Zou, Junchi Yan, Zuxuan Wu, and Yu-Gang Jiang.
\newblock Efficient-lvsm: Faster, cheaper and and better large view synthesis model via decoupled co-refinement attention.
\newblock \emph{arXiv preprint arXiv:2602.06478}, 2026.

\bibitem[Jiang et~al.(2025{\natexlab{a}})Jiang, Tan, Wang, Jin, Zhao, Bi, Zhang, Luan, Sunkavalli, Huang, and Pavlakos]{rayzer}
Hanwen Jiang, Hao Tan, Peng Wang, Haian Jin, Yue Zhao, Sai Bi, Kai Zhang, Fujun Luan, Kalyan Sunkavalli, Qixing Huang, and Georgios Pavlakos.
\newblock Rayzer: A self-supervised large view synthesis model.
\newblock In \emph{ICCV}, 2025{\natexlab{a}}.

\bibitem[Jiang et~al.(2025{\natexlab{b}})Jiang, Mao, Xu, Lu, Ren, Jin, Xu, Yu, Pang, Zhao, Lin, and Dai]{anysplat}
Lihan Jiang, Yucheng Mao, Linning Xu, Tao Lu, Kerui Ren, Yichen Jin, Xudong Xu, Mulin Yu, Jiangmiao Pang, Feng Zhao, Dahua Lin, and Bo~Dai.
\newblock Anysplat: Feed-forward 3d gaussian splatting from unconstrained views.
\newblock In \emph{SIGGRAPH ASIA}, 2025{\natexlab{b}}.

\bibitem[Jin et~al.(2025)Jin, Jiang, Tan, Zhang, Bi, Zhang, Luan, Snavely, and Xu]{lvsm}
Haian Jin, Hanwen Jiang, Hao Tan, Kai Zhang, Sai Bi, Tianyuan Zhang, Fujun Luan, Noah Snavely, and Zexiang Xu.
\newblock {LVSM:} {A} large view synthesis model with minimal 3d inductive bias.
\newblock In \emph{ICLR}, 2025.

\bibitem[Jin et~al.(2026)Jin, Wu, Zhang, Gao, Barron, Snavely, and Holynski]{zipmap}
Haian Jin, Rundi Wu, Tianyuan Zhang, Ruiqi Gao, Jonathan~T. Barron, Noah Snavely, and Aleksander Holynski.
\newblock Zipmap: Linear-time stateful 3d reconstruction via test-time training.
\newblock In \emph{CVPR}, 2026.

\bibitem[Johnson et~al.(2016)Johnson, Alahi, and Li]{perceptual}
Justin Johnson, Alexandre Alahi, and Fei-Fei Li.
\newblock Perceptual losses for real-time style transfer and super-resolution.
\newblock In \emph{ECCV}, 2016.

\bibitem[Jordan et~al.(2024)Jordan, Jin, Boza, You, Cesista, Newhouse, and Bernstein]{muon}
Keller Jordan, Yuchen Jin, Vlado Boza, Jiacheng You, Franz Cesista, Laker Newhouse, and Jeremy Bernstein.
\newblock Muon: An optimizer for hidden layers in neural networks, 2024.
\newblock URL \url{https://kellerjordan.github.io/posts/muon/}.

\bibitem[Kerbl et~al.(2023)Kerbl, Kopanas, Leimk{\"{u}}hler, and Drettakis]{3dgs}
Bernhard Kerbl, Georgios Kopanas, Thomas Leimk{\"{u}}hler, and George Drettakis.
\newblock 3d gaussian splatting for real-time radiance field rendering.
\newblock \emph{ACM TOG}, 2023.

\bibitem[Khakhulin et~al.(2022)Khakhulin, Korzhenkov, Solovev, Sterkin, Ardelean, and Lempitsky]{stereomag}
Taras Khakhulin, Denis Korzhenkov, Pavel Solovev, Gleb Sterkin, Andrei{-}Timotei Ardelean, and Victor Lempitsky.
\newblock Stereo magnification with multi-layer images.
\newblock In \emph{CVPR}, 2022.

\bibitem[Kim et~al.(2026)Kim, Ryu, Mitchel, and Sitzmann]{svsm}
Evan Kim, Hyunwoo Ryu, Thomas~W. Mitchel, and Vincent Sitzmann.
\newblock Scaling view synthesis transformers.
\newblock In \emph{CVPR}, 2026.

\bibitem[Liang et~al.(2025)Liang, Ren, Mirzaei, Torralba, Liu, Gilitschenski, Fidler, Oztireli, Ling, Gojcic, and Huang]{btimer}
Hanxue Liang, Jiawei Ren, Ashkan Mirzaei, Antonio Torralba, Ziwei Liu, Igor Gilitschenski, Sanja Fidler, Cengiz Oztireli, Huan Ling, Zan Gojcic, and Jiahui Huang.
\newblock Feed-forward bullet-time reconstruction of dynamic scenes from monocular videos.
\newblock \emph{Adv. Neural Inform. Process. Syst.}, 2025.

\bibitem[Lin et~al.(2025{\natexlab{a}})Lin, Sun, Yang, Chen, Lin, and Liu]{longsplat}
Chin{-}Yang Lin, Cheng Sun, Fu{-}En Yang, Min{-}Hung Chen, Yen{-}Yu Lin, and Yu{-}Lun Liu.
\newblock Longsplat: Robust unposed 3d gaussian splatting for casual long videos.
\newblock In \emph{ICCV}, 2025{\natexlab{a}}.

\bibitem[Lin et~al.(2025{\natexlab{b}})Lin, Wu, Yeh, Yen, Sun, and Liu]{lin2024frugalnerf}
Chin-Yang Lin, Chung-Ho Wu, Chang-Han Yeh, Shih-Han Yen, Cheng Sun, and Yu-Lun Liu.
\newblock Frugalnerf: Fast convergence for few-shot novel view synthesis without learned priors.
\newblock In \emph{CVPR}, 2025{\natexlab{b}}.

\bibitem[Ling et~al.(2024)Ling, Sheng, Tu, Zhao, Xin, Wan, Yu, Guo, Yu, Lu, Li, Sun, Ashok, Mukherjee, Kang, Kong, Hua, Zhang, Benes, and Bera]{dl3dv}
Lu~Ling, Yichen Sheng, Zhi Tu, Wentian Zhao, Cheng Xin, Kun Wan, Lantao Yu, Qianyu Guo, Zixun Yu, Yawen Lu, Xuanmao Li, Xingpeng Sun, Rohan Ashok, Aniruddha Mukherjee, Hao Kang, Xiangrui Kong, Gang Hua, Tianyi Zhang, Bedrich Benes, and Aniket Bera.
\newblock {DL3DV-10K:} {A} large-scale scene dataset for deep learning-based 3d vision.
\newblock In \emph{CVPR}, 2024.

\bibitem[Liu et~al.(2026)Liu, Elflein, Litany, Gojcic, and Li]{tttkvbind}
Junchen Liu, Sven Elflein, Or~Litany, Zan Gojcic, and Ruilong Li.
\newblock Test-time training with {KV} binding is secretly linear attention.
\newblock \emph{arxiv preprint arxiv:2602.21204}, 2026.

\bibitem[Liu et~al.(2022)Liu, Peng, Liu, Wang, Wang, Theobalt, Zhou, and Wang]{neuray}
Yuan Liu, Sida Peng, Lingjie Liu, Qianqian Wang, Peng Wang, Christian Theobalt, Xiaowei Zhou, and Wenping Wang.
\newblock Neural rays for occlusion-aware image-based rendering.
\newblock In \emph{CVPR}, 2022.

\bibitem[Loshchilov and Hutter(2017)]{adamw}
Ilya Loshchilov and Frank Hutter.
\newblock Decoupled weight decay regularization.
\newblock In \emph{ICLR}, 2017.

\bibitem[Meuleman et~al.(2023)Meuleman, Liu, Gao, Huang, Kim, Kim, and Kopf]{hike}
Andreas Meuleman, Y.~Liu, Chen Gao, Jia-Bin Huang, Changil Kim, Min~H. Kim, and Johannes Kopf.
\newblock Progressively optimized local radiance fields for robust view synthesis.
\newblock In \emph{CVPR}, 2023.

\bibitem[Mildenhall et~al.(2019)Mildenhall, Srinivasan, Cayon, Kalantari, Ramamoorthi, Ng, and Kar]{llff}
Ben Mildenhall, Pratul~P. Srinivasan, Rodrigo~Ortiz Cayon, Nima~Khademi Kalantari, Ravi Ramamoorthi, Ren Ng, and Abhishek Kar.
\newblock Local light field fusion: practical view synthesis with prescriptive sampling guidelines.
\newblock \emph{SIGGRAPH}, 2019.

\bibitem[Mildenhall et~al.(2020)Mildenhall, Srinivasan, Tancik, Barron, Ramamoorthi, and Ng]{nerf}
Ben Mildenhall, Pratul~P. Srinivasan, Matthew Tancik, Jonathan~T. Barron, Ravi Ramamoorthi, and Ren Ng.
\newblock Nerf: Representing scenes as neural radiance fields for view synthesis.
\newblock In \emph{ECCV}, 2020.

\bibitem[Mitchel et~al.(2026)Mitchel, Ryu, and Sitzmann]{truesn}
Thomas~W. Mitchel, Hyunwoo Ryu, and Vincent Sitzmann.
\newblock True self-supervised novel view synthesis is transferable.
\newblock In \emph{ICLR}, 2026.

\bibitem[M{\"{u}}ller et~al.(2022)M{\"{u}}ller, Evans, Schied, and Keller]{instant-ngp}
Thomas M{\"{u}}ller, Alex Evans, Christoph Schied, and Alexander Keller.
\newblock Instant neural graphics primitives with a multiresolution hash encoding.
\newblock \emph{ACM TOG}, 2022.

\bibitem[Plucker(1865)]{plucker}
Julius Plucker.
\newblock On a new geometry of space.
\newblock \emph{Philosophical Transactions of the Royal Society of London}, 1865.

\bibitem[Ren et~al.(2026)Ren, Tyszkiewicz, Huang, and Gojcic]{tokengs}
Jiawei Ren, Michal~Jan Tyszkiewicz, Jiahui Huang, and Zan Gojcic.
\newblock Tokengs: Decoupling 3d gaussian prediction from pixels with learnable tokens.
\newblock In \emph{CVPR}, 2026.

\bibitem[Rombach et~al.(2021)Rombach, Esser, and Ommer]{geofreevs}
Robin Rombach, Patrick Esser, and Bj{\"o}rn Ommer.
\newblock Geometry-free view synthesis: Transformers and no 3d priors.
\newblock In \emph{ICCV}, 2021.

\bibitem[Sajjadi et~al.(2022)Sajjadi, Meyer, Pot, Bergmann, Greff, Radwan, Vora, Lucic, Duckworth, Dosovitskiy, Uszkoreit, Funkhouser, and Tagliasacchi]{srt}
Mehdi S.~M. Sajjadi, Henning Meyer, Etienne Pot, Urs Bergmann, Klaus Greff, Noha Radwan, Suhani Vora, Mario Lucic, Daniel Duckworth, Alexey Dosovitskiy, Jakob Uszkoreit, Thomas~A. Funkhouser, and Andrea Tagliasacchi.
\newblock Scene representation transformer: Geometry-free novel view synthesis through set-latent scene representations.
\newblock In \emph{CVPR}, 2022.

\bibitem[Salimans and Kingma(2016)]{weightnormalization}
Tim Salimans and Diederik~P. Kingma.
\newblock Weight normalization: {A} simple reparameterization to accelerate training of deep neural networks.
\newblock In \emph{NeurIPS}, 2016.

\bibitem[Schlag et~al.(2021)Schlag, Irie, and Schmidhuber]{fastweight}
Imanol Schlag, Kazuki Irie, and J{\"{u}}rgen Schmidhuber.
\newblock Linear transformers are secretly fast weight programmers.
\newblock In \emph{ICML}, 2021.

\bibitem[Shazeer(2020)]{glu}
Noam Shazeer.
\newblock Glu variants improve transformer.
\newblock \emph{arxiv preprint arXiv:2002.05202}, 2020.

\bibitem[Shi et~al.(2016)Shi, Caballero, Husz{\'a}r, Totz, Aitken, Bishop, Rueckert, and Wang]{pixshufl}
Wenzhe Shi, Jose Caballero, Ferenc Husz{\'a}r, Johannes Totz, Andrew~P Aitken, Rob Bishop, Daniel Rueckert, and Zehan Wang.
\newblock Real-time single image and video super-resolution using an efficient sub-pixel convolutional neural network.
\newblock In \emph{CVPR}, 2016.

\bibitem[Sim'eoni et~al.(2025)Sim'eoni, Vo, Seitzer, Baldassarre, Oquab, Jose, Khalidov, Szafraniec, Yi, Ramamonjisoa, Massa, Haziza, Wehrstedt, Wang, Darcet, Moutakanni, Sentana, Roberts, Vedaldi, Tolan, Brandt, Couprie, Mairal, J'egou, Labatut, and Bojanowski]{dinov3}
Oriane Sim'eoni, Huy~V. Vo, Maximilian Seitzer, Federico Baldassarre, Maxime Oquab, Cijo Jose, Vasil Khalidov, Marc Szafraniec, Seungeun Yi, Michael Ramamonjisoa, Francisco Massa, Daniel Haziza, Luca Wehrstedt, Jianyuan Wang, Timoth{\'e}e Darcet, Th{\'e}o Moutakanni, Leonel Sentana, Claire Roberts, Andrea Vedaldi, Jamie Tolan, John Brandt, Camille Couprie, Julien Mairal, Herv'e J'egou, Patrick Labatut, and Piotr Bojanowski.
\newblock Dinov3.
\newblock \emph{arXiv preprint arXiv:2508.10104}, 2025.

\bibitem[Sitzmann et~al.(2021)Sitzmann, Rezchikov, Freeman, Tenenbaum, and Durand]{lfn}
Vincent Sitzmann, Semon Rezchikov, William~T. Freeman, Joshua~B. Tenenbaum, and Fr{\'e}do Durand.
\newblock Light field networks: Neural scene representations with single-evaluation rendering.
\newblock In \emph{Adv. Neural Inform. Process. Syst.}, 2021.

\bibitem[Suhail et~al.(2021)Suhail, Esteves, Sigal, and Makadia]{lfnr}
Mohammed Suhail, Carlos Esteves, Leonid Sigal, and Ameesh Makadia.
\newblock Light field neural rendering.
\newblock \emph{IEEE TPAMI}, 2021.

\bibitem[Suhail et~al.(2022)Suhail, Esteves, Sigal, and Makadia]{gpnr}
Mohammed Suhail, Carlos Esteves, Leonid Sigal, and Ameesh Makadia.
\newblock Generalizable patch-based neural rendering.
\newblock In \emph{ECCV}, 2022.

\bibitem[Sun et~al.(2022)Sun, Sun, and Chen]{dvgo}
Cheng Sun, Min Sun, and Hwann-Tzong Chen.
\newblock Direct voxel grid optimization: Super-fast convergence for radiance fields reconstruction.
\newblock In \emph{CVPR}, 2022.

\bibitem[Sun et~al.(2025{\natexlab{a}})Sun, Choe, Loop, Ma, and Wang]{svraster}
Cheng Sun, Jaesung Choe, Charles Loop, Wei{-}Chiu Ma, and Yu{-}Chiang~Frank Wang.
\newblock Sparse voxels rasterization: Real-time high-fidelity radiance field rendering.
\newblock In \emph{CVPR}, 2025{\natexlab{a}}.

\bibitem[Sun et~al.(2026)Sun, Choe, Chen, Hachiuma, and Wang]{dvsm}
Cheng Sun, Jaesung Choe, Min-Hung Chen, Ryo Hachiuma, and Yu-Chiang~Frank Wang.
\newblock Dvsm: Decoder-only view synthesis model done right.
\newblock \emph{arXiv preprint arXiv:2605.29891}, 2026.

\bibitem[Sun et~al.(2025{\natexlab{b}})Sun, Li, Dalal, Xu, Vikram, Zhang, Dubois, Chen, Wang, Koyejo, Hashimoto, and Guestrin]{ttt}
Yu~Sun, Xinhao Li, Karan Dalal, Jiarui Xu, Arjun Vikram, Ge~Zhang, Yann Dubois, Xinlei Chen, Xiaolong Wang, Oluwasanmi Koyejo, Tatsunori Hashimoto, and Carlos Guestrin.
\newblock Learning to (learn at test time): Rnns with expressive hidden states.
\newblock In \emph{ICML}, 2025{\natexlab{b}}.

\bibitem[Szymanowicz et~al.(2024)Szymanowicz, Insafutdinov, Zheng, Campbell, Henriques, Rupprecht, and Vedaldi]{flash3d}
Stanislaw Szymanowicz, Eldar Insafutdinov, Chuanxia Zheng, Dylan Campbell, Jo{\~a}o~F. Henriques, Christian Rupprecht, and Andrea Vedaldi.
\newblock Flash3d: Feed-forward generalisable 3d scene reconstruction from a single image.
\newblock \emph{3DV}, 2024.

\bibitem[Szymanowicz et~al.(2026)Szymanowicz, Chen, Wang, Rupprecht, and Vedaldi]{largernvs}
Stanislaw Szymanowicz, Minghao Chen, Jianyuan Wang, Christian Rupprecht, and Andrea Vedaldi.
\newblock Lagernvs: Latent geometry for fully neural real-time novel view synthesis.
\newblock In \emph{CVPR}, 2026.

\bibitem[Tang et~al.(2024)Tang, Chen, Chen, Wang, Zeng, and Liu]{lgm}
Jiaxiang Tang, Zhaoxi Chen, Xiaokang Chen, Tengfei Wang, Gang Zeng, and Ziwei Liu.
\newblock Lgm: Large multi-view gaussian model for high-resolution 3d content creation.
\newblock In \emph{ECCV}, 2024.

\bibitem[Wang et~al.(2026{\natexlab{a}})Wang, Tan, Yifan, Chen, Liu, Sunkavalli, Bi, Liu, and Hu]{tttlrm}
Chen Wang, Hao Tan, Wang Yifan, Zhiqin Chen, Yuheng Liu, Kalyan Sunkavalli, Sai Bi, Lingjie Liu, and Yiwei Hu.
\newblock tttlrm: Test-time training for long context and autoregressive 3d reconstruction.
\newblock \emph{arXiv preprint arXiv:2602.20160}, 2026{\natexlab{a}}.

\bibitem[Wang et~al.(2026{\natexlab{b}})Wang, Ye, Li, Chen, and Chen]{thely}
Haoru Wang, Kai Ye, Yangyan Li, Wenzheng Chen, and Baoquan Chen.
\newblock The less you depend, the more you learn: Synthesizing novel views from sparse, unposed images without any 3d knowledge.
\newblock In \emph{ICLR}, 2026{\natexlab{b}}.

\bibitem[Wang et~al.(2025{\natexlab{a}})Wang, Chen, Karaev, Vedaldi, Rupprecht, and Novotn{\'{y}}]{vggt}
Jianyuan Wang, Minghao Chen, Nikita Karaev, Andrea Vedaldi, Christian Rupprecht, and David Novotn{\'{y}}.
\newblock {VGGT:} visual geometry grounded transformer.
\newblock In \emph{CVPR}, 2025{\natexlab{a}}.

\bibitem[Wang et~al.(2023)Wang, Liu, Chen, Liu, Liu, Komura, Theobalt, and Wang]{f2nerf}
Peng Wang, Yuan Liu, Zhaoxi Chen, Lingjie Liu, Ziwei Liu, Taku Komura, Christian Theobalt, and Wenping Wang.
\newblock F2-nerf: Fast neural radiance field training with free camera trajectories.
\newblock In \emph{CVPR}, 2023.

\bibitem[Wang et~al.(2021)Wang, Wang, Genova, Srinivasan, Zhou, Barron, Martin{-}Brualla, Snavely, and Funkhouser]{ibrnet}
Qianqian Wang, Zhicheng Wang, Kyle Genova, Pratul~P. Srinivasan, Howard Zhou, Jonathan~T. Barron, Ricardo Martin{-}Brualla, Noah Snavely, and Thomas~A. Funkhouser.
\newblock Ibrnet: Learning multi-view image-based rendering.
\newblock In \emph{CVPR}, 2021.

\bibitem[Wang et~al.(2024)Wang, Chen, Wang, and Huang]{pmsnet}
Xingzheng Wang, Kaiqiang Chen, Zixuan Wang, and Wenhao Huang.
\newblock Pmsnet: Parallel multi-scale network for accurate low-light light-field image enhancement.
\newblock In \emph{Transactions on Multimedia}, 2024.

\bibitem[Wang et~al.(2025{\natexlab{b}})Wang, Wu, Chen, Zhang, and Furukawa]{clift}
Zhengqing Wang, Yuefan Wu, Jiacheng Chen, Fuyang Zhang, and Yasutaka Furukawa.
\newblock Clift: Compressive light-field tokens for compute-efficient and adaptive neural rendering.
\newblock In \emph{NeurIPS}, 2025{\natexlab{b}}.

\bibitem[Wei et~al.(2024)Wei, Zhang, Bi, Tan, Luan, Deschaintre, Sunkavalli, Su, and Xu]{meshlrm}
Xinyue Wei, Kai Zhang, Sai Bi, Hao Tan, Fujun Luan, Valentin Deschaintre, Kalyan Sunkavalli, Hao Su, and Zexiang Xu.
\newblock Meshlrm: Large reconstruction model for high-quality mesh.
\newblock \emph{arXiv preprint arXiv:2404.12385}, 2024.

\bibitem[Xu et~al.(2025)Xu, Peng, Wang, Blum, Barath, Geiger, and Pollefeys]{depthsplat}
Haofei Xu, Songyou Peng, Fangjinhua Wang, Hermann Blum, Daniel Barath, Andreas Geiger, and Marc Pollefeys.
\newblock Depthsplat: Connecting gaussian splatting and depth.
\newblock In \emph{CVPR}, 2025.

\bibitem[Yang et~al.(2020)Yang, Mao, {\'A}lvarez, and Liu]{cvpmvsnet}
Jiayu Yang, Wei Mao, Jos{\'e}~Manuel {\'A}lvarez, and Miaomiao Liu.
\newblock Cost volume pyramid based depth inference for multi-view stereo.
\newblock In \emph{CVPR}, 2020.

\bibitem[Yao et~al.(2018)Yao, Luo, Li, Fang, and Quan]{mvsnet}
Yao Yao, Zixin Luo, Shiwei Li, Tian Fang, and Long Quan.
\newblock Mvsnet: Depth inference for unstructured multi-view stereo.
\newblock In \emph{ECCV}, 2018.

\bibitem[Yu et~al.(2021)Yu, Ye, Tancik, and Kanazawa]{pixelnerf}
Alex Yu, Vickie Ye, Matthew Tancik, and Angjoo Kanazawa.
\newblock pixelnerf: Neural radiance fields from one or few images.
\newblock In \emph{CVPR}, 2021.

\bibitem[Yu et~al.(2024)Yu, Xing, Yuan, Hu, Li, Huang, Gao, Wong, Shan, and Tian]{viewcrafter}
Wangbo Yu, Jinbo Xing, Li~Yuan, Wenbo Hu, Xiaoyu Li, Zhipeng Huang, Xiangjun Gao, Tien-Tsin Wong, Ying Shan, and Yonghong Tian.
\newblock Viewcrafter: Taming video diffusion models for high-fidelity novel view synthesis.
\newblock \emph{IEEE TPAMI}, 2024.

\bibitem[Zhang et~al.(2024)Zhang, Bi, Tan, Xiangli, Zhao, Sunkavalli, and Xu]{gslrm}
Kai Zhang, Sai Bi, Hao Tan, Yuanbo Xiangli, Nanxuan Zhao, Kalyan Sunkavalli, and Zexiang Xu.
\newblock {GS-LRM:} large reconstruction model for 3d gaussian splatting.
\newblock In \emph{ECCV}, 2024.

\bibitem[Zhang et~al.(2025)Zhang, Bi, Hong, Zhang, Luan, Yang, Sunkavalli, Freeman, and Tan]{lact}
Tianyuan Zhang, Sai Bi, Yicong Hong, Kai Zhang, Fujun Luan, Songlin Yang, Kalyan Sunkavalli, William~T Freeman, and Hao Tan.
\newblock Test-time training done right.
\newblock \emph{arXiv preprint arXiv:2505.23884}, 2025.

\bibitem[Zhao et~al.(2025)Zhao, Tan, Wang, Bi, Zhang, Sunkavalli, Tulsiani, and Jiang]{erayzer}
Qitao Zhao, Hao Tan, Qianqian Wang, Sai Bi, Kai Zhang, Kalyan Sunkavalli, Shubham Tulsiani, and Hanwen Jiang.
\newblock E-rayzer: Self-supervised 3d reconstruction as spatial visual pre-training.
\newblock \emph{arXiv preprint arXiv:2512.10950}, 2025.

\bibitem[Zhou et~al.(2025)Zhou, Gao, Voleti, Vasishta, Yao, Boss, Torr, Rupprecht, and Jampani]{seva}
Jensen Zhou, Hang Gao, Vikram~S. Voleti, Aaryaman Vasishta, Chun-Han Yao, Mark Boss, Philip Torr, Christian Rupprecht, and Varun Jampani.
\newblock Stable virtual camera: Generative view synthesis with diffusion models.
\newblock In \emph{ICCV}, 2025.

\end{thebibliography}
